\documentclass[10pt,twocolumn,letterpaper]{article}

\usepackage[pagenumbers]{cvpr} % To force page numbers, e.g. for an arXiv version

\usepackage{times}
\usepackage{epsfig}
\usepackage{graphicx}
\usepackage{amsmath}
\usepackage{amssymb}
\usepackage{booktabs}
\usepackage{color}
\usepackage{caption}
\usepackage{subcaption}
\usepackage{makecell}
\usepackage{multirow}

\usepackage{bbding}
\usepackage{url}

\usepackage{float}
\usepackage{stfloats}
\usepackage{balance}

\usepackage{tcolorbox}
\tcbuselibrary{breakable,skins}
\usepackage{xcolor}
\usepackage[table]{xcolor}

\usepackage{circledsteps}

\usepackage[utf8]{inputenc}
\usepackage{pifont}

\usepackage{rotating}

\definecolor{cvprblue}{rgb}{0.21,0.49,0.74}
\definecolor{salmon}{RGB}{255,103,125}

\definecolor{textblue}{RGB}{221,235,247}
\definecolor{audiogreen}{RGB}{226,239,218}
\definecolor{videoyellow}{RGB}{255,242,204}
\definecolor{trajpurple}{RGB}{234,209,220}
\definecolor{rhythmred}{RGB}{244,204,204}

\usepackage[pagebackref,breaklinks,colorlinks,allcolors=cvprblue]{hyperref}

\newcommand{\cmark}{\textcolor{blue!80!black}{\scalebox{1.2}{\ding{51}}}}
\newcommand{\xmark}{\textcolor{red!80!black}{\scalebox{1.2}{\ding{55}}}}

\newcommand{\best}[1]{\textbf{#1}}
\newcommand{\second}[1]{\underline{#1}}

\newcommand{\bgbs}[2]{#1/#2}
\newcommand{\bgbsfmt}[2]{#1/#2}

\usepackage{longtable}
\usepackage{array}
\usepackage{ragged2e}
\usepackage{xurl}
\usepackage{microtype}

\usepackage{hyperref}
\usepackage{tabularx}

\def\paperID{****} % *** Enter the Paper ID here
\def\confName{CVPR}
\def\confYear{2026}

\title{Benchmarking Behavioral Steerability in Behavior Foundation Models}

\author{
Minghe Gao\textsuperscript{1,2,3} \thanks{Work done in Joint Laboratory of Embodied Intelligence, ZJU \& Unitree Robotics}, 
Zhanxi Yan\textsuperscript{1},
Jiahui Liu\textsuperscript{1,4},
Wendong Bu\textsuperscript{1},
Xiaoting Chen\textsuperscript{1}, \\
Qizhou Wang\textsuperscript{2}\footnotemark[1],
Yi Su\textsuperscript{2},
Siliang Tang\textsuperscript{1}, 
Jun Xiao\textsuperscript{1}, 
Yueting Zhuang\textsuperscript{1},
Tat-Seng Chua\textsuperscript{3},
Juncheng Li\textsuperscript{1}\thanks{Corresponding author.}
\\
{\small \textsuperscript{1}ZJU \quad \textsuperscript{2}Unitree \quad \textsuperscript{3}NUS \quad \textsuperscript{4}CSU } 
\\
{\url{https://robosteer.github.io/}}
}

\begin{document}
\maketitle

\begin{abstract}

Behavior Foundation Models (BFMs) are emerging as a paradigm for translating human intentions into executable humanoid behaviors. As these models evolve beyond behavior generation toward general-purpose behavioral systems, a fundamental question arises: can they be reliably steered according to user intentions? In this paper, we introduce the concept of \textit{behavioral steerability}, defined as the ability of BFMs to faithfully generate behaviors that satisfy user-specified intentions. To study this capability, we present \textbf{\textit{RoboSteer}}, the first benchmark for behavioral steerability in BFMs. \textbf{\textit{RoboSteer}} organizes behavioral steerability into a three-level hierarchy—Conditional Steering, Constraint Steering, and Compositional Steering—and establishes a unified evaluation framework supported by a large-scale multimodal motion corpus. Using \textbf{\textit{RoboSteer}}, we conduct the first large-scale empirical study of behavioral steerability across 9 existing BFMs. We view behavioral steerability as more than a capability for controlling motion: it concerns how embodied systems translate human intentions into purposeful actions. We hope \textbf{\textit{RoboSteer}} will advance research on intention realization as a foundation for general-purpose embodied intelligence.

\end{abstract}

\section{Introduction}

Behavior Foundation Models (BFMs) represent an emerging family of humanoid foundation models that directly translate diverse human intentions into executable whole-body behaviors through a single general-purpose model~\cite{ICLR2025_37611b0f,zeng2025behavior,yuan2025survey,li2026bfm,bjorck2025gr00t,zeng2026scaling}. Unlike task-specific robot policies or vision-language-action models that predict low-level control actions, BFMs operate in a unified behavior space, enabling motion generation, imitation, completion, and editing under diverse multimodal conditions~\cite{xie2026textop,bian2025motioncraft,jiang2025uniact}. These capabilities have enabled increasingly sophisticated humanoid behaviors. However, generating complex behaviors does not mean that a model can reliably satisfy specific user requirements. As BFMs become a more general interface for human--robot interaction, an important question is how to adapt their behavior generation to diverse and evolving human intentions.

\begin{figure}
    \centering
    \includegraphics[width=1\linewidth]{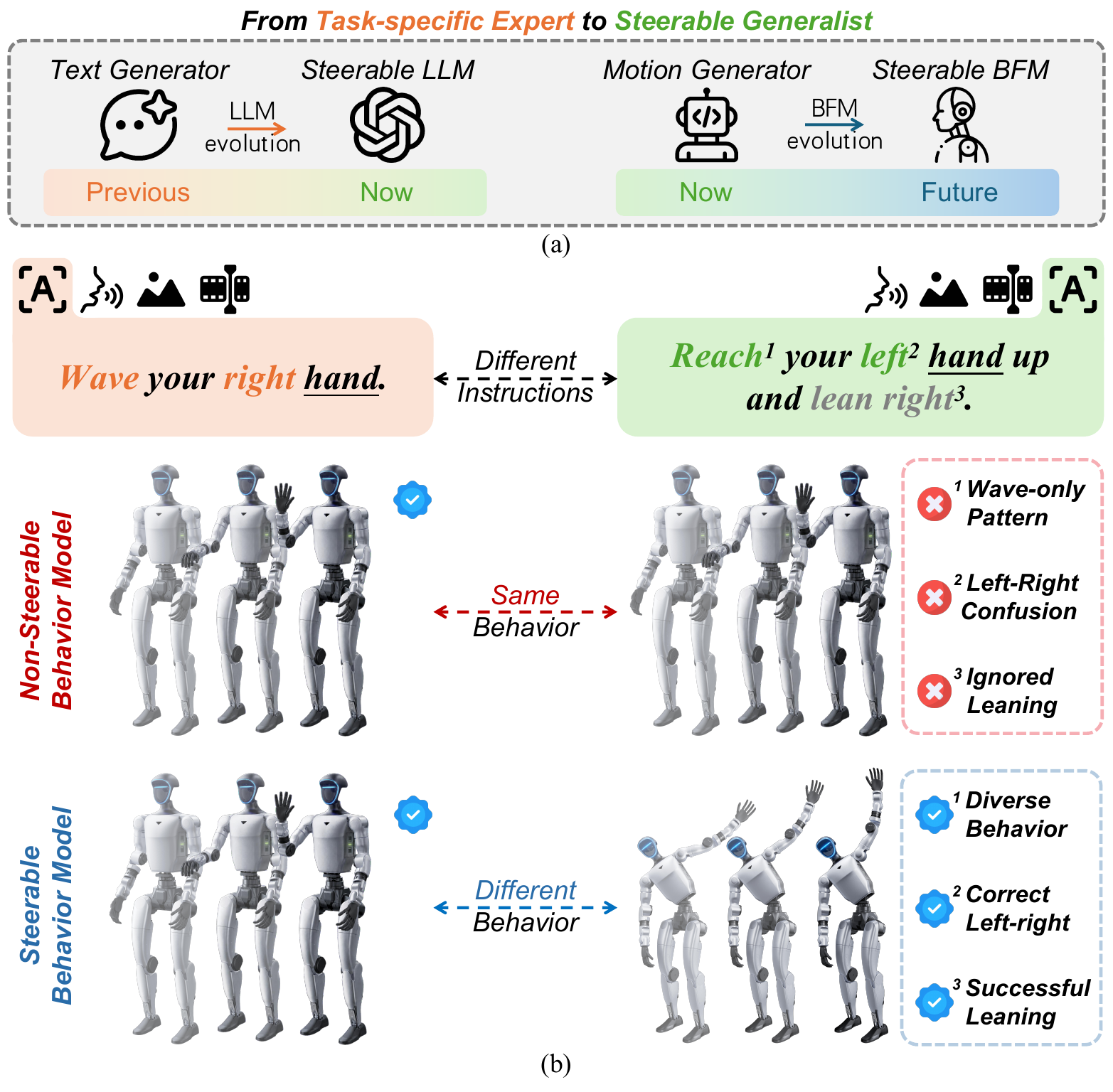}
    \vspace{-5mm}
    \caption{(a) LLM evolution provides precedent for steerable foundation models. (b) Steerable vs. non-steerable behavior models.}
    \label{fig1}
    \vspace{-5mm}
\end{figure}

The evolution of large language models offers a precedent. As shown in Figure~\ref{fig1}\textcolor{cvprblue}{(a)}, as language foundation models became increasingly general, the emergence of ChatGPT~\cite{openai2022chatgpt} transformed them from powerful text generators into steerable interactive systems~\cite{achiam2023gpt,yang2025qwen3,liu2024deepseek}. This transition redefined the role of foundation models: beyond generating high-quality outputs, they are now expected to be reliably steered according to human intentions. BFMs are now entering a similar stage: as they become the primary interface through which humans communicate behavioral intentions to robots, their role extends beyond synthesizing realistic behaviors to reliably translating diverse human intentions into the intended behaviors. This raises a fundamental question:

\begin{center}
\textit{\textbf{Can a behavior foundation model be steerable?}}
\end{center}

In this paper, we define \textbf{behavioral steerability} as the ability of BFMs to generate behaviors that faithfully satisfy user intentions. Moving from behavior generators to general behavioral systems requires more than generating physically plausible behaviors; BFMs must also reliably adapt behavior generation to diverse user requirements, including desired conditions, execution constraints, or compositional instructions. In real-world human--robot interaction, users rarely request unconstrained behavior~\cite{karunratanakul2023guided,xie2024omnicontrol,yuan2023physdiff}; instead, they specify how a behavior should be performed. For example, a user may ask a robot to follow a reference motion while moving faster, avoiding one arm, and passing through a target location. Such requests remain challenging for existing BFMs, which are primarily optimized to generate behaviors rather than adapting them to evolving user intentions~\cite{tevet2022human,zhang2024motiondiffuse,guo2022generating}. We therefore argue that behavioral steerability is a defining capability to evolve from behavior generators into general behavioral systems. Figure~\ref{fig1}\textcolor{cvprblue}{(b)} illustrates the distinction.

\begin{figure*}[t]
\centering

% 左边：表格
\begin{subfigure}[b]{0.6\textwidth}
    \vspace{0pt}
    \centering

    \resizebox{0.98\linewidth}{!}{
        \renewcommand{\arraystretch}{1.5}

        \begin{tabular}{@{}l|c|c|lll|ll|c@{}}
        \toprule

        \textbf{Benchmark}
        & \textbf{Platform}
        & \makecell{\textbf{Evaluation}\\\textbf{Target}}
        & \multicolumn{3}{c|}{\makecell{\textbf{Behavioral}\\ \textbf{Steering}}}
        & \multicolumn{2}{c|}{\textbf{Task}}
        & \multirow{2}{*}{\rotatebox{90}{\textbf{Number of modality}}}
        \\

        \cline{4-8}
        &
        &
        &
        \rotatebox{90}{\textbf{Conditional}}
        &
        \rotatebox{90}{\textbf{Constraint}}
        &
        \rotatebox{90}{\textbf{Compositional  }}
        &
        \rotatebox{90}{\textbf{Temporal}}
        &
        \rotatebox{90}{\textbf{Spatial}}
        &
        \\

        \midrule

        HumanoidBench
        & Humanoid
        & Task / Skill
        & \xmark & \xmark & \xmark & \xmark & \xmark & 1
        \\

        Mimicking-Bench
        & Humanoid
        & Skill / Motion
        & \xmark & \xmark & \xmark & \xmark & \cmark & 2
        \\

        HumanoidArena
        & Humanoid
        & Task / Skill
        & \xmark & \xmark & \xmark & \xmark & \xmark & 2
        \\

        Humanoid Everyday
        & Humanoid
        & Task / Skill
        & \xmark & \xmark & \xmark & \xmark & \xmark & 2
        \\

        Meta-World
        & Manipulation
        & Task
        & \xmark & \xmark & \xmark & \xmark & \xmark & 1
        \\

        LIBERO
        & Manipulation
        & Task / Skill
        & \xmark & \xmark & \xmark & \xmark & \xmark & 2
        \\

        \midrule

        \textbf{RoboSteer (Ours)}
        & \textbf{Humanoid}
        & \textbf{Behavior}
        & \cmark & \cmark & \cmark & \cmark & \cmark & \textbf{5}
        \\

        \bottomrule
        \end{tabular}
    }
\caption{ }
\label{fig2a}
\end{subfigure}
\hfill
% 右边：图片
\begin{subfigure}[b]{0.37\textwidth}
\vspace{0pt}
\centering
\includegraphics[width=\linewidth]{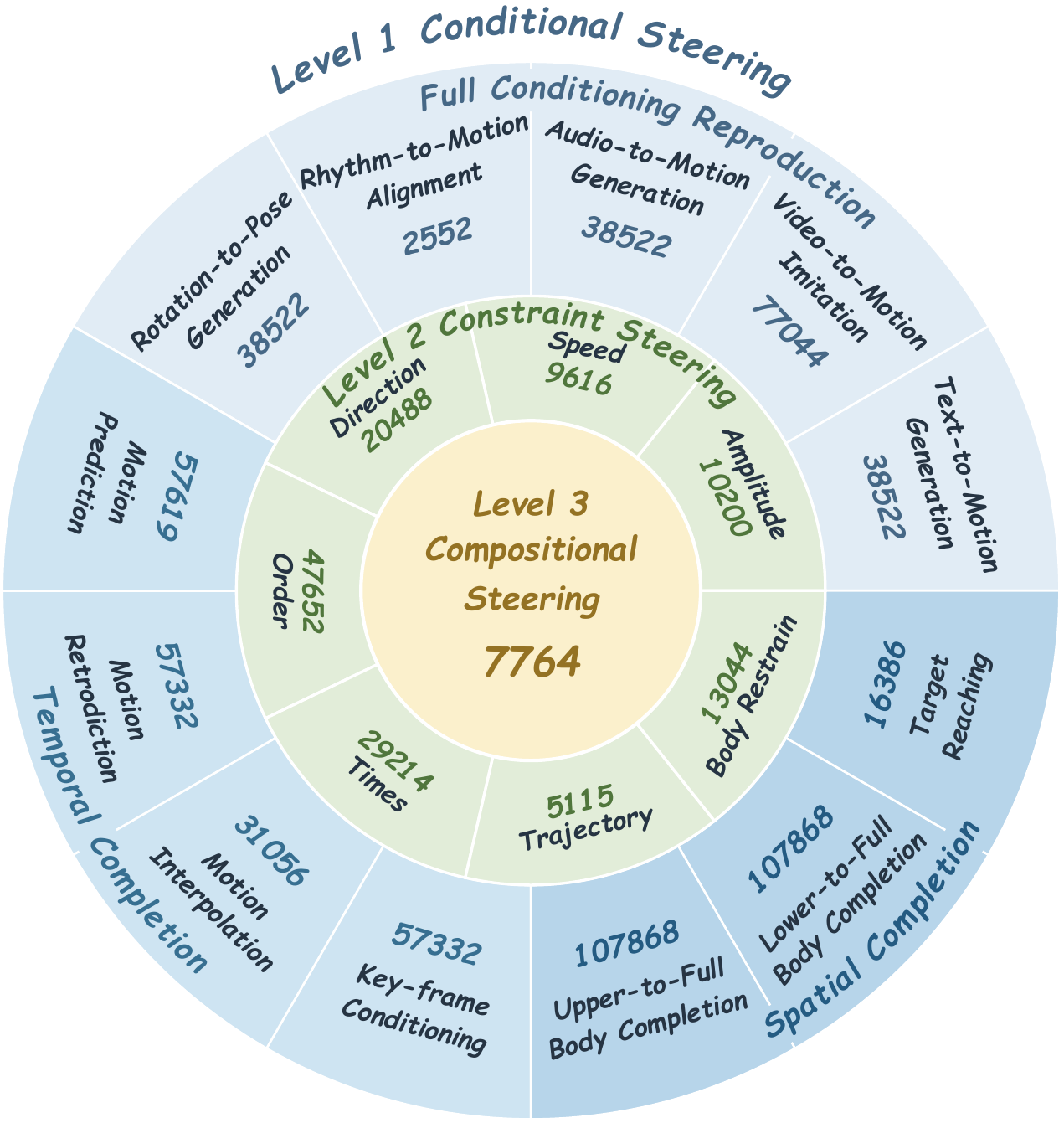}
\caption{ }
\label{fig2b}
\end{subfigure}
\vspace{-3mm}
\caption{(a) Comparison with existing benchmarks. (b) Dataset statistics of the \textbf{\textit{RoboSteer}}.}
\vspace{-3mm}
\end{figure*}

To study behavioral steerability, we introduce \textbf{\textit{RoboSteer}}, the first benchmark for defining and evaluating this capability in BFMs. As shown in Figure~\ref{fig3}, \textbf{\textit{RoboSteer}} organizes behavioral steerability into three levels: \textbf{Conditional Steering}, which evaluates behavior generation under a specified behavioral condition (e.g., following a reference motion or a textual behavior description); \textbf{Constraint Steering}, which measures compliance with explicit behavioral requirements (e.g., movement speed, body-part restrictions, or target locations); and \textbf{Compositional Steering}, which evaluates whether models can simultaneously satisfy multiple interleaved behavioral requirements from different sources (e.g., imitating a reference video and then following a textual instruction with a speed constraint). \textbf{\textit{RoboSteer}} further offers an open-source framework for evaluating behavioral steerability in simulation and real-world settings.

To support this benchmark, we leverage pre-processed web-scale human motion data and retarget and track the motions to form a multimodal corpus for humanoid behavioral steerability evaluation. The data are represented as joint-level motion trajectories, together with associated videos, text, audio, and keyframes. The resulting corpus comprises 38,522 source motions, 77,044 human and rendered sequences, totaling 91.58 hours and 773,716 tasks.

Using \textbf{\textit{RoboSteer}}, we conduct an empirical study of behavioral steerability across 9 BFMs with diverse paradigms and multimodal inputs. Our results reveal a clear gap between behavior generation and behavioral steerability, highlighting the challenges introduced by increasingly complex steering requirements. Current BFMs also exhibit weaker fine-grained spatial and body-part control, while struggling to satisfy multiple behavioral requirements simultaneously. These findings highlight behavioral steerability as a distinct capability beyond behavior generation and suggest the need for more fine-grained and multimodal behavioral control.

Our findings invite a broader question: what does it mean for embodied intelligence to move beyond generating behaviors toward realizing human intentions? The ability to fulfill intentions gives behaviors their purpose in human--robot interaction, enabling robots to adapt what they do and how they do it to human needs. By formalizing behavioral steerability, \textbf{\textit{RoboSteer}} provides a starting point for investigating this capability as a potential foundation for more general embodied intelligence. Its implications may extend from intention-guided low-level control to higher-level perception, decision-making, and action in VLA and VLN systems. Understanding how intention realization can shape these capabilities may open new directions toward embodied intelligence that is not only capable of acting, but also responsive to human goals, advancing toward the broader vision of AGI.

\section{Existing Related Works}

\subsection{Behavior Generation as Dominant Paradigm}

Existing research on BFMs has largely been organized around behavior generation. Recent advances span motion generation, imitation learning, motion completion, motion editing, multimodal behavior modeling, and embodied interaction, all aiming to improve the ability of BFMs to generate realistic and executable behaviors~\cite{yuan2025survey,zeng2025behavior,pirotta2024fast,jiang2023motiongpt,zhang2023t2mgpt}. Correspondingly, numerous datasets and benchmarks have been developed to evaluate behavior generation capability through measures such as motion quality, generation fidelity, and task completion~\cite{Guo_2022_CVPR,punnakkal2021babel,NEURIPS2023_4f8e27f6}. Together, these efforts implicitly assume that successful behavior generation provides a sufficient characterization of BFM capability. As BFMs continue to evolve toward more general behavioral systems, however, generating high-quality behaviors alone is no longer sufficient; models must also be evaluated by how faithfully they transform human intentions into executable behaviors.

\subsection{Toward Steerable Behavior Systems}
Recent research has increasingly explored controllable behavior generation as BFMs become more general and interactive. Representative efforts include motion editing, instruction-conditioned behavior generation, constraint-aware motion synthesis, and vision-language-action models that condition behaviors on language or multimodal observations~\cite{tevet2022motionclip,xie2024omnicontrol,ai2026pi07}. Collectively, these studies signal an important shift in humanoid behavior modeling—from generating behaviors to controlling how behaviors are generated. However, these approaches primarily propose new control mechanisms rather than defining behavior control as a measurable capability that can be systematically compared across BFMs. Consequently, the field still lacks a unified formulation and evaluation framework for behavioral steerability.

\begin{figure}
    \centering
    \includegraphics[width=1\linewidth]{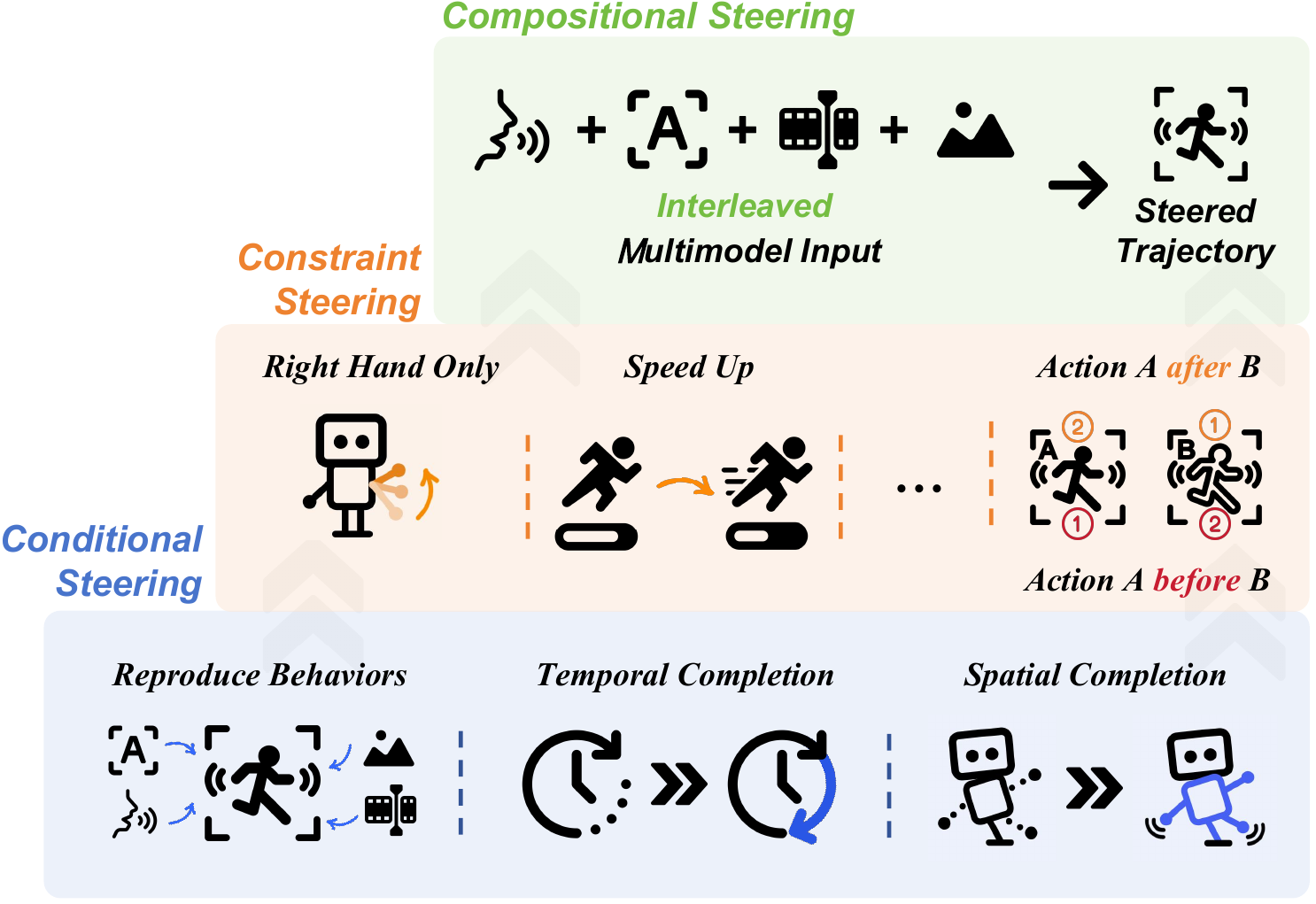}
    \vspace{-5mm}
    \caption{Behavioral steerability is built upon behavior generation and progressively evolves from satisfying simple to complex behavioral specifications.}
    \vspace{-3mm}
    \label{fig3}
\end{figure}

\begin{figure*}
    \centering
    \includegraphics[width=0.95\linewidth]{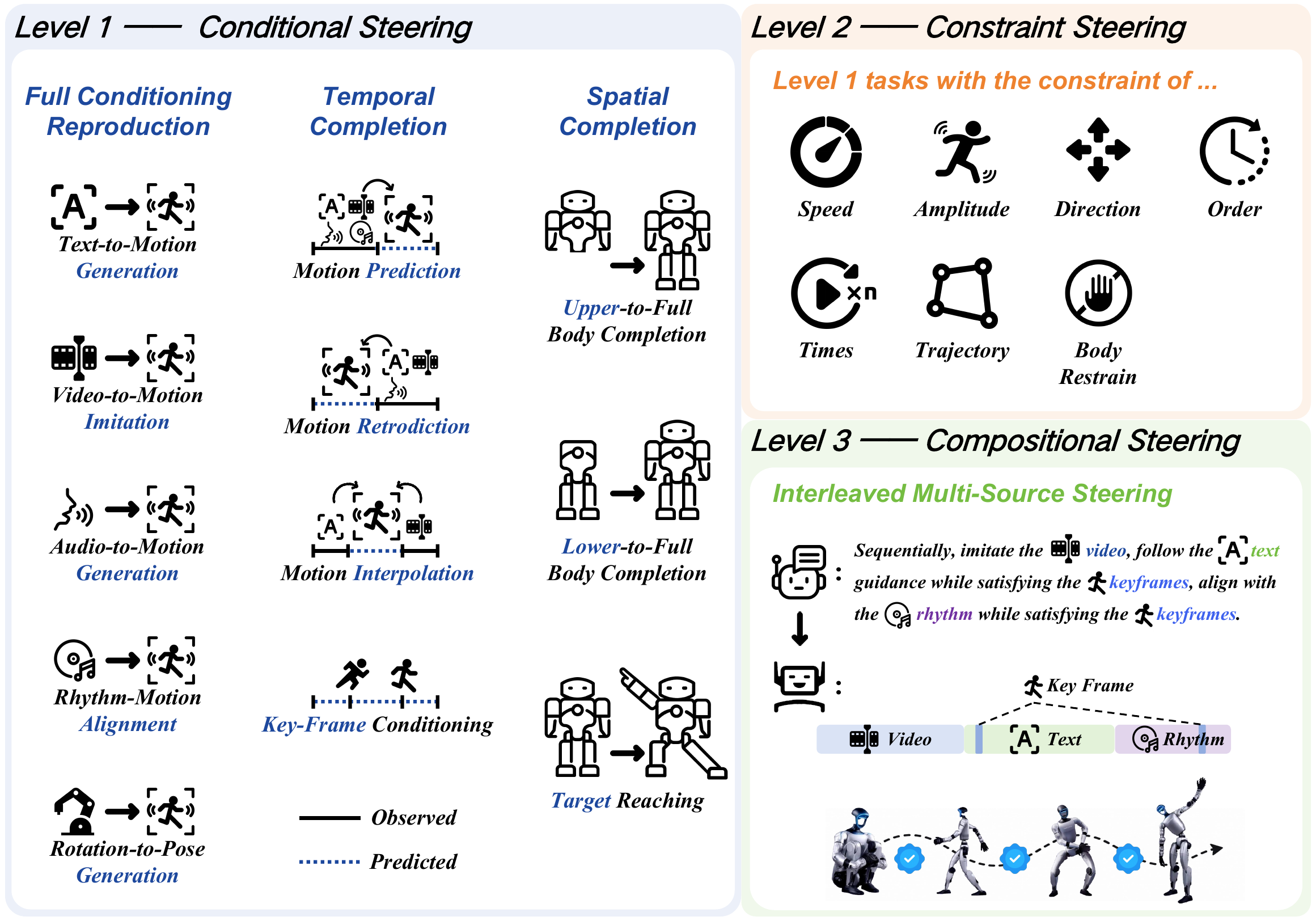}
    \caption{Overview of the three-level behavioral steerability hierarchy in \textbf{\textit{RoboSteer}}.}
    \vspace{-3mm}
    \label{fig4}
\end{figure*}

\section{Behavioral Steerability}

\noindent\textbf{Formulation.} Existing behavior generation paradigms can generally be formulated as
\begin{equation}
B=\arg\max_{B}P(B\mid x),
\end{equation}
\text{or}
\begin{equation}
B=f_{\theta}(x),
\end{equation}
where $x$ denotes the behavioral input and $B$ denotes the resulting behavior. These formulations establish a mapping from input conditions to behaviors, typically within a behavioral control space defined by the capabilities and patterns supported by the model. Even when the input accepts free-form descriptions, the resulting behavior is often grounded in previously learned behavioral patterns, which limits the ability to realize behavioral requirements beyond the supported space. We instead characterize \textbf{behavioral steerability} by considering the intention. Let
\begin{equation}
B = f_\theta(x), \qquad
\max \mathcal{F}\bigl(B, I(x)\bigr),
\tag{3}
\end{equation}
where $\mathcal{I}(x)$ denotes the behavioral intention conveyed by the input $x$, and $\mathcal{F}\bigl(B,\mathcal{I}(x)\bigr)$ characterizes the degree to which the generated behavior $B$ realizes this intention. Under this formulation, behavioral steerability does not require a model to explicitly represent or extract the intention. Instead, it characterizes the extent to which the behavior generated from $x$ aligns with and realizes the behavioral intention conveyed by $x$. Behavioral steerability therefore extends behavior generation from producing a behavior appropriate to the input to maximizing the degree of intention realization, which can be quantified through task-specific intention realization measures at different levels of behavioral steering.

\noindent\textbf{Concept.} Behavioral steerability is defined on top of successful behavior generation. A model must first generate an executable behavior that satisfies the underlying behavioral objective before its ability to faithfully realize the behavioral intention can be meaningfully evaluated. For example, if a humanoid robot is instructed to raise its right leg but instead loses balance or produces an invalid motion, the failure occurs at the level of behavior generation, regardless of whether the behavior was correctly understood. Behavioral steerability therefore preserves the objectives of behavior generation while introducing an additional intention-alignment objective, requiring the generated behavior to be not only appropriate to the input, but also faithful to the behavioral intention.

\noindent\textbf{Characteristics.}
Behavioral steerability is not a binary capability, but one that naturally exhibits progressively increasing complexity as behavioral specifications become increasingly sophisticated. At its simplest, user intentions describe a desired behavior under a single behavioral specification. More realistic scenarios further introduce explicit execution specifications that regulate how the behavior should be performed, such as movement speed, body-part usage, or target locations. In practical human--robot interaction, users often specify multiple heterogeneous behavioral requirements simultaneously, requiring behavior models to satisfy several behavioral specifications while preserving the intended behavior. These diverse forms of behavioral specification naturally give rise to progressively more demanding levels of behavioral steerability, as illustrated in Figure~\ref{fig3}.

\section{RoboSteer}

The formulation and characteristics of behavioral steerability naturally motivate a systematic benchmark. We present \textbf{\textit{RoboSteer}}, the first benchmark for behavioral steerability in BFMs. The remainder of this section introduces its definition (Section~\ref{sec4.1}), construction (Section~\ref{sec4.2}), evaluation protocol (Section~\ref{sec4.3}), and analysis (Section~\ref{sec4.4}), providing a unified framework for evaluating behavioral steerability.

\subsection{Task Definition}
\label{sec4.1}

As illustrated in Figure~\ref{fig4}, \textbf{\textit{RoboSteer}} organizes behavioral steerability into three challenging levels. The three levels characterize a progression from realizing behaviors under a single behavioral condition, to satisfying explicit requirements on how a behavior should be performed, and ultimately to jointly satisfying multiple interleaved requirements from heterogeneous sources. For completeness, we provide one representative example for every single task in the \textcolor{red}{Homepage}.

\noindent \textbf{\textit{Level 1: Conditional Steering}.}
It examines whether a model can realize a behavior under conditions with different degrees of spatiotemporal completeness. It covers three forms of conditioning: \textit{Full Conditioning Reproduction}, where the behavioral condition is fully specified through modalities such as text, video, audio, rhythm, or rotation-to-pose cues; \textit{Temporal Completion}, where only a partial temporal segment is provided and the remaining behavior must be completed through motion prediction, retrodiction, interpolation, or key-frame conditioning; and \textit{Spatial Completion}, where only part of the spatial or body configuration is specified and the missing components must be completed, including upper- and lower-to-full-body completion, and target reaching. These tasks establish the basic steering capability of realizing both fully specified behaviors and behaviors that require temporal or spatial completion from partial conditions.

\noindent \textbf{\textit{Level 2: Constraint Steering}.}
It builds upon \textit{Level 1} by addressing a more practical requirement: realizing a desired behavior is often insufficient when users also need to specify \emph{how} it should be performed. We therefore consider seven constraints: \textit{Speed}, \textit{Amplitude}, \textit{Direction}, \textit{Order}, \textit{Times}, \textit{Trajectory}, and \textit{Body Restrain}. Each \textit{Level 2} task extends a corresponding \textit{Level 1} task with one semantic constraint, thereby evaluating whether a model can preserve the intended behavior while satisfying an additional requirement.

\noindent \textbf{\textit{Level 3: Compositional Steering}.}
It further requires models to jointly satisfy multiple interleaved behavioral requirements from heterogeneous sources. We formulate it as \textit{Interleaved Multi-Source Steering}, where instructions from sources such as video, text, audio, and keyframes must be sequentially followed within a single behavior. For example, a model may be required to imitate a reference video, follow textual guidance, align with a rhythmic pattern, and satisfy key-frame requirements in order. This level focuses on the coordinated realization of multiple temporally interleaved behavioral requirements across heterogeneous modalities.

\subsection{Construction}
\label{sec4.2}

We construct \textbf{\textit{RoboSteer}} from web-scale human motion videos through an automatic video-to-humanoid-motion pipeline, which performs quality filtering, semantic annotation, human motion reconstruction, and physically constrained humanoid retargeting. A simulation-based tracking policy further verifies the executability of reconstructed motions, retaining only motions that can be stably tracked by the humanoid robot. Based on the resulting motion data and metadata, we construct the three benchmark levels by completing task-specific annotations and transforming the motions into the corresponding conditional, constrained, and compositional settings. Specifically, \textit{Level~1} is instantiated from diverse conditional generation and completion tasks, \textit{Level~2} augments the corresponding \textit{Level~1} tasks with explicit behavioral constraints, and \textit{Level~3} composes multimodal inputs into temporally interleaved instructions. Detailed data processing, annotation, and task construction procedures are provided in the \textcolor{red}{Appendix}~\ref{Appendix_B}.

\begin{table}[t]
\centering
\small
\setlength{\tabcolsep}{5pt}
\renewcommand{\arraystretch}{1.15}
\resizebox{\linewidth}{!}{%
\begin{tabular}{@{}l|l|l@{}}
\toprule
\textbf{Metric}
& \textbf{What it measures}
& \textbf{Comparison / Target}
\\
\midrule

\textbf{FID} $\downarrow$
& Generated motion quality
& $B$ vs. $B_{GT}$
\\

\textbf{Diversity} $\uparrow$
& Motion diversity
& Generated behaviors
\\

\textbf{Contact Sliding} $\downarrow$
& Physical plausibility
& $B$
\\

\textbf{MM-Distance} $\downarrow$
& Behavior--instruction consistency
& $B$ vs. $x$
\\

\textbf{R@1} $\uparrow$
& Behavior--instruction correspondence
& $B$ vs. $x$
\\

\textbf{MPJPE} $\downarrow$
& Motion reconstruction accuracy
& $B$ vs. $B_{GT}$
\\

\textbf{$E_{\mathrm{vel}}$} $\downarrow$
& Motion velocity consistency
& $B$ vs. $B_{GT}$
\\

\textbf{BAS-Gen} $\uparrow$
& Rhythm-conditioned generation
& $B$ vs. rhythm condition
\\

\textbf{BAS-Gap} $\downarrow$
& Rhythm consistency
& $B$ vs. rhythm condition
\\

\midrule

\textbf{BG} $\uparrow$
& Unified behavior generation score
& $B$--$B_{GT}$ and $B$--$x$
\\

\textbf{IR\textsubscript{$k$}} $\uparrow$
& Intention realization at Level $k$
& $B$--$\mathcal{I}_k(x)$
\\

\textbf{BS\textsubscript{$l$}} $\uparrow$
& Cumulative behavior steering score
& $\mathrm{BG}\times\prod_{k=1}^{l}\mathrm{IR}_k$
\\

\bottomrule
\end{tabular}
}
\caption{
Summary of conventional generation metrics and the proposed BG and BS scores.
}
\label{tab:metric_summary}
\end{table}

\begin{figure}
    \centering
    \includegraphics[width=1\linewidth]{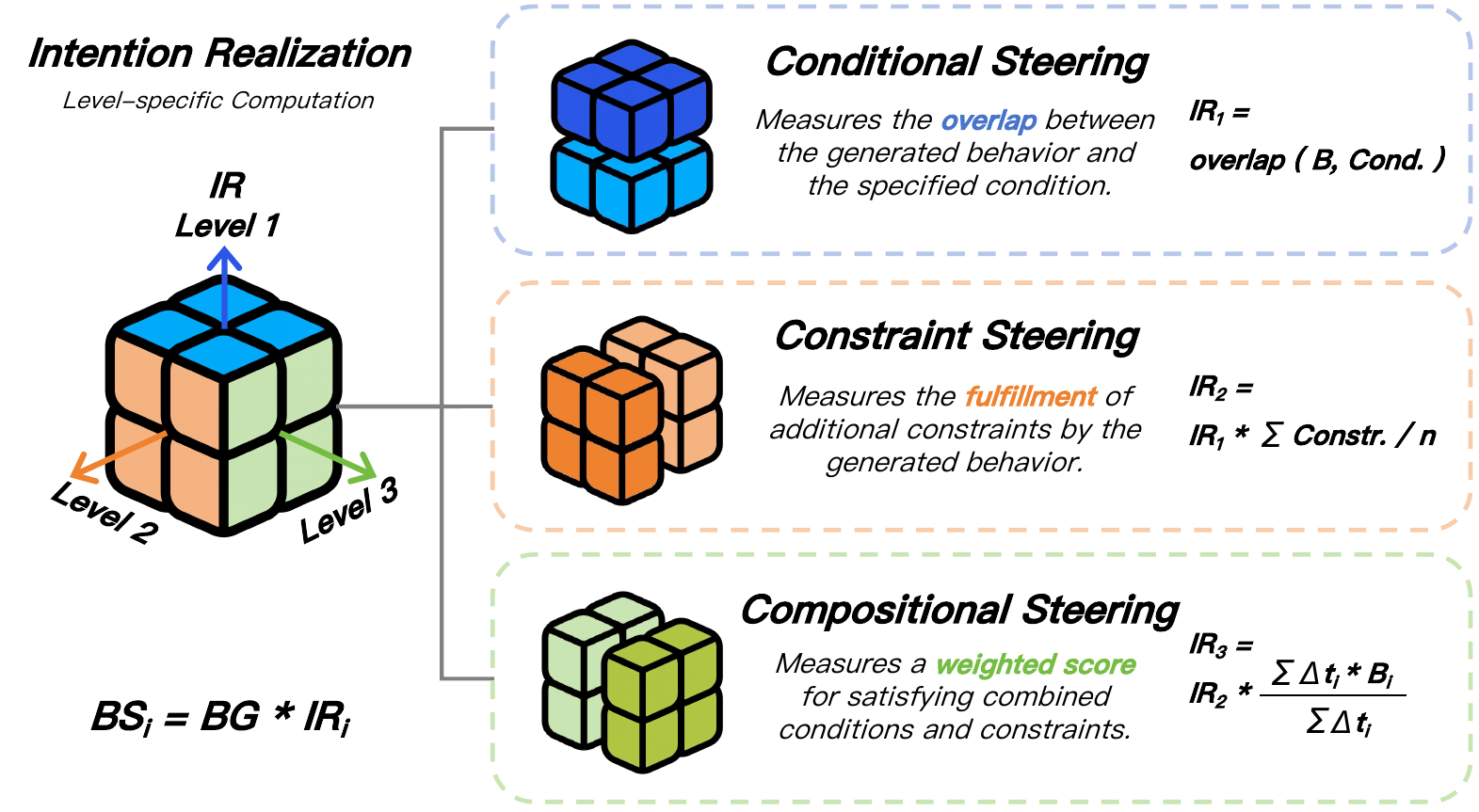}
    \caption{The computation of Intention Realization across three steering levels.}
    \label{fig5}
\end{figure}

\subsection{Evaluation Matrix}
\label{sec4.3}

\noindent\textbf{\textit{Conventional Evaluation Matrix.}}
We first establish a conventional evaluation matrix using metrics widely adopted in existing behavior generation methods. These metrics assess different aspects of generated behaviors, including overall motion quality (\textit{FID}), motion diversity (\textit{Diversity}), physical plausibility (\textit{Contact Sliding}), consistency with the input condition (\textit{MM-Distance}, \textit{R@1}), motion reconstruction accuracy (\textit{MPJPE}), velocity consistency (\textit{$E_{\mathrm{vel}}$}), and rhythm generation and consistency (\textit{BAS-Gen}, \textit{BAS-Gap}). Their definitions and evaluation targets are summarized in Table~\ref{tab:metric_summary}. Although these metrics provide complementary evaluations of behavior generation, they remain largely individual, with each metric focusing on a particular property or relationship. Consequently, conventional evaluation does not provide a unified score that jointly captures these complementary aspects of behavior generation.

\noindent\textbf{\textit{Behavior Generation Score.}}
To unify these complementary aspects, we introduce the Behavior Generation Score (BG), which jointly accounts for the consistency of the generated behavior with both the target behavior and the input instruction. Specifically, we use \textbf{\textit{FID}} and \textbf{\textit{MM-Distance}} as representative metrics for these two relationships and normalize them such that higher values indicate better performance. BG is then defined as
\[
\mathrm{BG} = 1000 \cdot \exp\left(-0.3 \cdot \mathrm{FID} - 1.6 \cdot \mathrm{MM\text{-}Distance}\right)
\]
We empirically determine the coefficients based on experiments to balance the two metrics and prevent either from dominating the BG score. The multiplicative formulation requires both aspects to be satisfied simultaneously, such that poor consistency with either the target behavior or the input directly reduces BG. However, BG only evaluates consistency with the target behavior and input, rather than whether the generated behavior satisfies the intended behavioral requirements. Thus, BG alone is insufficient for evaluating behavioral steerability, which requires assessing additional behavioral requirements beyond basic behavior generation.

\begin{table*}[t]
\centering
\small

% Uniform inter-column padding.
\setlength{\tabcolsep}{2.5pt}
\renewcommand{\arraystretch}{1.15}

\resizebox{\textwidth}{!}{%
\begin{tabular}{
l|
c<{\hspace{8pt}}cc|
cc|
c|
ccc
}
\toprule

% =========================
% Top-level header
% =========================
\multicolumn{1}{l|}{\textbf{Model}}
& \multicolumn{6}{c|}{\textbf{Generation Task (Level 1 Reproduction Part)}}
& \multicolumn{3}{c}{\textbf{Steering Task}}
\\

\cmidrule{1-1}
\cmidrule(lr){2-7}
\cmidrule(lr){8-10}

% =========================
% Text
% =========================
\rowcolor{textblue}
\textbf{Text}
& \textbf{FID} $\downarrow$
& \textbf{Diversity} $\uparrow$
& \textbf{Contact Sliding} $\downarrow$
& \textbf{MM-Distance} $\downarrow$
& \textbf{R@1/R@5} $\uparrow$
& \textbf{BG} $\uparrow$
& \textbf{BS\textsubscript{1}} $\uparrow$
& \textbf{BS\textsubscript{2}} $\uparrow$
& \textbf{BS\textsubscript{3}} $\uparrow$
\\

GEM + GMR
& \underline{0.387}
& \underline{0.786}
& \underline{0.530}
& \textbf{1.284}
& 0.043/0.198
& \underline{114.094}
& \textbf{68.375}
& 16.148
& \textbf{42.122}
\\

LoM + GMR
& 0.586
& 0.598
& \textbf{0.380}
& \underline{1.303}
& 0.036/0.173
& 104.633
& \underline{66.266}
& 16.825
& 39.276
\\

MotionCraft + GMR
& \textbf{0.208}
& \textbf{0.895}
& 0.541
& 1.314
& \textbf{0.061/0.262}
& \textbf{114.928}
& 63.981
& \textbf{22.734}
& \underline{40.659}
\\

TextOp
& 0.812
& 0.690
& 7.620
& 1.408
& 0.035/0.168
& 82.673
& 43.539
& \underline{20.407}
& 31.053
\\

UniAct
& 0.439
& 0.729
& 0.647
& 1.343
& 0.041/0.186
& 102.444
& 57.952
& 12.755
& 37.052
\\

UH-1
& 0.768
& 0.678
& 0.633
& 1.323
& \underline{0.045/0.209}
& 96.647
& 48.970
& 18.529
& 38.208
\\

\midrule

% =========================
% Audio
% =========================
\rowcolor{audiogreen}
\textbf{Audio}
& \textbf{FID} $\downarrow$
& \textbf{Diversity} $\uparrow$
& \textbf{Contact Sliding} $\downarrow$
& \textbf{MM-Distance} $\downarrow$
& \textbf{R@1/R@5} $\uparrow$
& \textbf{BG} $\uparrow$
& \textbf{BS\textsubscript{1}} $\uparrow$
& \textbf{BS\textsubscript{2}} $\uparrow$
& \textbf{BS\textsubscript{3}} $\uparrow$
\\

GEM + GMR
& \textbf{0.387}
& \underline{0.786}
& \underline{0.530}
& 1.363
& 0.039/0.179
& 100.556
& \underline{59.751}
& 14.826
& \underline{39.930}
\\

LoM + GMR
& 0.586
& 0.598
& \textbf{0.380}
& \textbf{1.314}
& 0.035/0.168
& \underline{102.507}
& \textbf{65.062}
& 16.586
& 38.007
\\

MotionCraft + GMR
& \underline{0.390}
& \textbf{0.817}
& 0.868
& \underline{1.320}
& \textbf{0.052/0.223}
& \textbf{108.233}
& 57.875
& \underline{19.744}
& \textbf{40.047}
\\

TextOp
& 0.783
& 0.679
& 7.599
& 1.321
& 0.033/0.166
& 95.589
& 51.305
& \textbf{22.366}
& 34.434
\\

UniAct
& 0.470
& 0.669
& 0.662
& 1.343
& 0.038/0.181
& 101.369
& 54.570
& 12.891
& 36.513
\\

UH-1
& 0.768
& 0.678
& 0.632
& 1.333
& \underline{0.042/0.192}
& 94.796
& 48.500
& 18.300
& 36.615
\\

\midrule

% =========================
% Video
% =========================
\rowcolor{videoyellow}
\textbf{Video}
& \textbf{FID} $\downarrow$
& \textbf{Diversity} $\uparrow$
& \textbf{Contact Sliding} $\downarrow$
& \makecell{\textbf{MPJPE/g-MPJPE} $\downarrow$}
& \textbf{E\textsubscript{vel}} $\downarrow$
& \textbf{BG} $\uparrow$
& \textbf{BS\textsubscript{1}} $\uparrow$
& \textbf{BS\textsubscript{2}} $\uparrow$
& \textbf{BS\textsubscript{3}} $\uparrow$
\\

GEM + GMR
& \underline{0.290}
& \underline{0.956}
& \underline{1.488}
& \underline{293.187}/\textbf{624.422}
& \textbf{16.673}
& \underline{114.876}
& \underline{78.494}
& \underline{20.978}
& \textbf{36.573}
\\

Video2Robot + GMR
& \textbf{0.262}
& \textbf{1.031}
& \textbf{0.730}
& \textbf{228.112}/\underline{3046.974}
& \underline{17.658}
& \textbf{119.344}
& \textbf{91.984}
& \textbf{23.730}
& \underline{34.973}
\\

\midrule

% =========================
% Trajectory
% =========================
\rowcolor{trajpurple}
\textbf{Trajectory}
& \textbf{FID} $\downarrow$
& \textbf{Diversity} $\uparrow$
& \textbf{Contact Sliding} $\downarrow$
& \makecell{\textbf{MPJPE/g-MPJPE} $\downarrow$}
& \textbf{E\textsubscript{vel}} $\downarrow$
& \textbf{BG} $\uparrow$
& \textbf{BS\textsubscript{1}} $\uparrow$
& \textbf{BS\textsubscript{2}} $\uparrow$
& \textbf{BS\textsubscript{3}} $\uparrow$
\\

BFM-Zero
& \textbf{0.839}
& \textbf{1.249}
& \textbf{0.275}
& \textbf{198.973/339.253}
& \textbf{11.404}
& \textbf{95.387}
& \textbf{95.387}
& -
& -
\\

\midrule

% =========================
% Rhythm
% =========================
\rowcolor{rhythmred}
\textbf{Rhythm}
& \textbf{FID} $\downarrow$
& \textbf{Diversity} $\uparrow$
& \textbf{Contact Sliding} $\downarrow$
& \textbf{BAS-Gen} $\uparrow$
& \textbf{BAS-Gap} $\downarrow$
& \textbf{BG} $\uparrow$
& \textbf{BS\textsubscript{1}} $\uparrow$
& \textbf{BS\textsubscript{2}} $\uparrow$
& \textbf{BS\textsubscript{3}} $\uparrow$
\\

LDA + GMR
& \underline{1.016}
& \underline{0.451}
& \textbf{0.487}
& \textbf{0.443}
& \underline{0.165}
& \underline{94.009}
& \underline{94.009}
& -
& -
\\

MotionCraft
& \textbf{0.754}
& \textbf{0.586}
& \underline{2.928}
& \underline{0.436}
& \textbf{0.158}
& \textbf{96.045}
& \textbf{96.045}
& -
& -
\\

\bottomrule
\end{tabular}
}

\caption{
Evaluation results on \textbf{\textit{RoboSteer}} across different modalities and progressively increasing levels of behavioral steerability.
}
\label{tab1}
\end{table*}

\noindent\textbf{\textit{Behavior Steering Score.}}
To address this limitation, we introduce the Behavior Steering Score (BS) by extending BG with an \textit{Intention Realization} score:
\[
\mathrm{BS}_{l}
=
\mathrm{BG}
\prod_{k=1}^{l}\mathrm{IR}_{k},
\qquad l\in\{1,2,3\}.
\]
where $\mathrm{IR}_{k}\in[0,1]$ measures the extent to which the behavioral intention introduced at steering level $k$ is successfully realized. Unlike conventional generation metrics that directly evaluate model outputs, IR is outcome-oriented and is derived from feedback during robot execution, including the robot state and its interaction with the environment. It therefore evaluates whether the intended behavioral requirement is actually realized during execution, rather than how well the generated motion itself matches a reference. The three levels of \textbf{\textit{RoboSteer}} capture distinct forms of behavioral steering: \textit{Conditional Steering}, \textit{Constraint Steering}, and \textit{Compositional Steering}. Each level introduces a different type of steering requirement, and the corresponding intention realization factor is defined according to the task structure at that level. BS retains the behavior generation capability measured by BG while further evaluating whether the generated behavior can satisfy increasingly complex behavioral intentions. The specific computation of $\mathrm{IR}_{k}$ depends on the behavioral requirements and task structure at each level, while detailed task-level definitions can be found at Figure~\ref{fig5} and computation procedures are provided in the \textcolor{red}{Appendix}~\ref{Appendix_C}.

\subsection{Dataset Statistics}
\label{sec4.4}

\textbf{\textit{RoboSteer}} comprises over \textbf{77,044} human and rendered motion sequences spanning \textbf{91.58} hours and \textbf{773,716} behavioral tasks, providing large-scale coverage for evaluating behavioral steerability in BFMs. Beyond dataset scale, Figure~\ref{fig2a} compares \textbf{\textit{RoboSteer}} with existing behavior benchmarks across multiple dimensions, highlighting its broader evaluation coverage. As summarized in Figure~\ref{fig2b}, the dataset spans the three steering levels, multiple behavioral modalities, seven explicit constraint types, and temporally interleaved multi-source instructions, covering both temporal- and spatial-aware behavior control. Compared with existing humanoid behavior benchmarks, \textbf{\textit{RoboSteer}} therefore provides broader coverage of the behavioral conditions, constraints, and compositional instructions required for systematic evaluation of behavioral steerability. Detailed statistics and distributions are provided in the \textcolor{red}{Appendix}~\ref{Appendix_A}.

\section{Experiment}

\subsection{Experimental Setup}

\textbf{Model Setup.}
We organize evaluated models by their supported conditioning modalities, including text, audio, video, rhythm, and trajectory. As the existing models produce motions in either human-centric or robot-centric spaces, we retarget human-centric motions to the target humanoid embodiment and evaluate all outputs in a unified robot-centric space. We further perform simulation-based motion tracking to validate the physical executability of generated motions. Whenever available, we use official checkpoints, pretrained weights, demo models, or released inference packages to ensure faithful reproduction. We will release the official adaptation and evaluation code.

\textbf{Baselines.}
We evaluate nine representative model families under their supported input modalities, with the detailed settings summarized in Table~\ref{tab1}. For text-conditioned tasks, we evaluate GENMO~\cite{gem2025}, LOM~\cite{languageofmotion2025}, MotionCraft~\cite{motioncraft2025}, TextOP~\cite{xie2026textop}, UniAct~\cite{uniact2025}, and UH-1~\cite{uh1_2025}. For audio-conditioned tasks, speech inputs are first transcribed into text using Whisper, after which we apply the same six text-conditioned baselines. For video-conditioned tasks, we evaluate GENMO~\cite{gem2025} and PromptHMR~\cite{prompthmr2025}. For trajectory-conditioned tasks, we evaluate BFM-Zero~\cite{bfmzero2026}. For rhythm-conditioned tasks, we evaluate LDA~\cite{lda2023} and MotionCraft~\cite{motioncraft2025}. Human-centric motions are retargeted to the robot embodiment using GMR~\cite{gmr2026}, a general human-to-humanoid motion-retargeting framework, while simulation-based validation is conducted with SONIC~\cite{luo2026sonic}, an open-source humanoid whole-body motion-tracking framework. Whenever available, we use official checkpoints, pretrained weights, demo models, or released inference packages to ensure faithful reproduction. The prompts, settings, and hyperparameters used for each model are detailed in the \textcolor{red}{Appendix}~\ref{Appendix_D}.

\subsection{Main Results and Analysis}

The main results are summarized in Table~\ref{tab1}, with complete results provided in the \textcolor{red}{Appendix}~\ref{Appendix_E}. We analyze the results through six questions that examine the gap between behavior generation and steerability, the effect of steering complexity, the role of conditioning modalities, the adequacy of conventional evaluation metrics, and the challenges introduced by partial conditions and explicit behavioral constraints.

\noindent\textbf{Q1: Does behavioral capability imply behavioral steerability?}
No. Across evaluated models, conventional generation metrics and \textbf{BG} scores remain relatively competitive, indicating that current BFMs have acquired a reasonable capability to generate valid and instruction-relevant behaviors. However, this capability does not directly translate into behavioral steerability. The corresponding \textbf{BS} scores are consistently lower than BG, with the gap becoming particularly pronounced when additional steering requirements are introduced. This discrepancy demonstrates that generating a behavior and reliably modifying that behavior according to user requirements constitute distinct capabilities. In other words, strong behavior-generation performance alone is insufficient to guarantee effective behavioral steering.

\noindent\textbf{Q2: How does steerability change with increasing behavioral complexity?}
Behavioral steerability varies with behavioral complexity, with more complex steering scenarios introducing additional challenges for current BFMs. While individual behavioral requirements can often be effectively translated into motion, coordinating multiple requirements becomes more challenging as the behavioral specification becomes richer. In particular, complex steering requires models to simultaneously interpret, integrate, and execute heterogeneous behavioral requirements, revealing limitations in maintaining consistent and precise control throughout the motion. These results suggest that the key challenge extends beyond behavior generation itself to the effective coordination of multiple behavioral requirements during execution.

\begin{figure*}[t]
    \centering
    \includegraphics[width=0.95\linewidth]{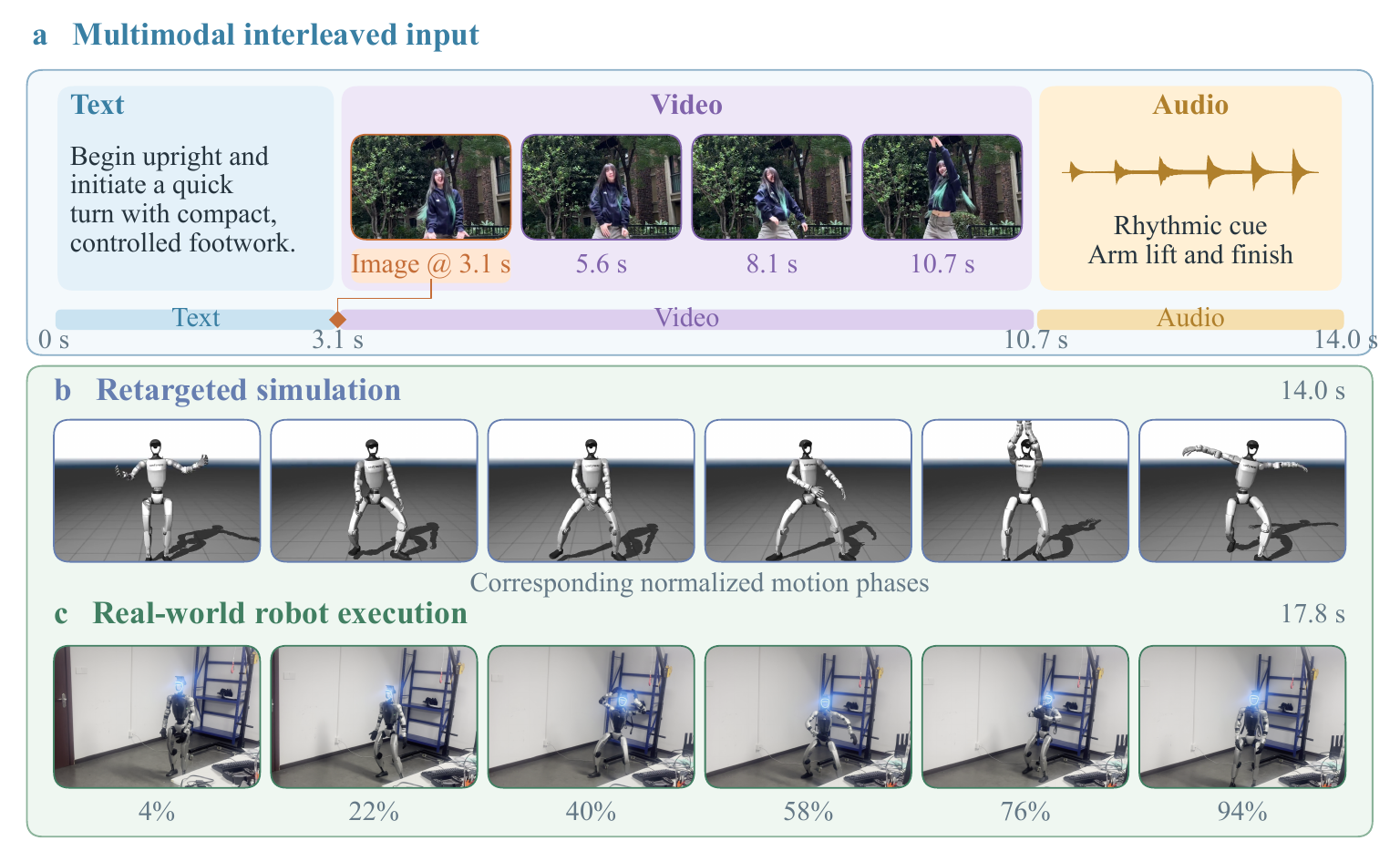}
    \vspace{-3mm}
    \caption{Real-world demonstration of a \textit{Compositional Steering} task and its model execution. More tasks are provided in the \textcolor{red}{Homepage}.}
    \vspace{-3mm}
    \label{fig6}
\end{figure*}

\noindent\textbf{Q3: Do current BFMs exhibit balanced temporal and spatial steerability?}
According to the Level~1 results detailed in Appendix Table~\ref{tab:level1_tasks}, spatial completion achieves lower $\mathrm{BS}_{\mathrm{Level~1}}$ scores than temporal completion across most models, indicating that current BFMs exhibit weaker steerability in the spatial dimension. We hypothesize that this gap is partly rooted in the prevailing pre-training paradigm of motion models, which predominantly emphasizes predicting how behaviors evolve over time, while placing less emphasis on modeling whole-body motion from the perspective of an embodied agent. In particular, such models may not sufficiently capture the fine-grained spatial dependencies among individual body joints that are essential for maintaining coordinated and physically coherent whole-body movements. This limitation suggests that future BFMs should go beyond modeling temporal dynamics and explicitly incorporate fine-grained joint-level spatial representations and their coordination into pre-training.

\noindent\textbf{Q4: Which behavioral constraints are most challenging for current BFMs?}
According to the Level~2 results detailed in Appendix Table~~\ref{tab:level2_tasks}, \textit{Body Restrain} is the most challenging constraint across models. Unlike constraints that primarily modify global motion properties, body-specific constraints require the model to establish a correspondence between the instruction and the targeted body part while maintaining balance. We hypothesize that current BFMs are limited in this regard because their pre-training typically does not explicitly encourage localized modeling or control of individual joints and body regions. As a result, the models may capture holistic motion patterns without developing fine-grained representations that associate behavioral instructions with specific parts of the body. This observation further corroborates the finding in Q3 that current BFMs exhibit weaker spatial steerability, highlighting the importance of joint- and body-region-level representations for precise whole-body behavioral control.

\noindent\textbf{Q5: What capabilities are required for future BFMs to achieve reliable compositional steering?}
According to the Level~3 results, current BFMs struggle to reliably satisfy multiple behavioral requirements simultaneously. We argue that this limitation is partly rooted in the predominantly language-centric paradigm of existing behavioral models. Complex physical behaviors often contain information that is difficult to precisely express through language alone, such as fine-grained spatial configurations, motion patterns, and body-part relationships. As a result, relying on a single modality can create an inherent bottleneck in specifying compositional behavioral intentions. Future BFMs should therefore move beyond language-only behavioral specification toward multimodal, interleaved, and potentially streaming inputs, where different modalities can provide complementary information about the desired behavior. Such multimodal behavioral interfaces could enable richer specification of complex behavioral intentions, providing an important foundation for reliable compositional behavioral steering.

\noindent\textbf{Q6: What do the results reveal about the current state and future of behavioral steering?} Our results suggest that behavioral steering remains a fundamentally open problem. \textbf{\textit{RoboSteer}} progressively examines temporal and spatial requirements at Level~1, multiple execution constraints at Level~2, and interleaved multimodal requirements at Level~3, yet current BFMs still struggle to consistently translate such requirements into precise and coordinated behavior. Despite rapid progress in behavior foundation models, a mature and general-purpose BFM with the broad behavioral steerability of language models such as ChatGPT remains elusive. We therefore view \textbf{\textit{behavioral steerability}} as an important capability and quantitative dimension for future BFMs, rather than a solved problem. Through \textbf{\textit{RoboSteer}}, we hope to provide an initial foundation for measuring this capability and to continuously maintain and extend the benchmark as the field progresses.

\section{Conclusion}

\textbf{\textit{RoboSteer}} is introduced as a comprehensive benchmark for studying behavioral steerability in BFMs. Through extensive evaluation, we reveal that behavior generation does not necessarily translate into behavioral steerability, highlighting the need to explicitly evaluate how reliably models can realize user-specified intentions. As the project progressed, we increasingly recognized that general-purpose embodied intelligence extends beyond steerability itself, encompassing broader challenges such as interaction, simulation, and closed-loop feedback. Nevertheless, we believe that steerability, as represented by \textbf{\textit{RoboSteer}}, can serve as an important starting point for tackling these broader challenges and paving the way toward more comprehensive, general, and intelligent embodied systems.

\clearpage

{
    \small
    \bibliographystyle{ieeenat_fullname}
    \bibliography{main}
}

\appendix
\onecolumn

\renewcommand{\thetable}{B.\arabic{table}}
\setcounter{table}{0}

\setlength{\LTpre}{0.6em}
\setlength{\LTpost}{0.8em}
\setlength{\LTleft}{0pt}
\setlength{\LTright}{0pt}
\renewcommand{\arraystretch}{1.16}

\emergencystretch=3em

\newcommand{\taskheading}[1]{%
  \par\addvspace{0.8\baselineskip}%
  \noindent{\large\bfseries #1\par}%
  \nobreak\addvspace{0.25\baselineskip}%
}

\newtcolorbox{promptbox}[1][]{
  enhanced,
  breakable,
  colback=gray!8,          % 背景浅灰
  colframe=black!80,       % 边框颜色
  boxrule=0.8pt,           % 边框粗细
  arc=4mm,                 % 圆角大小
  left=6mm,
  right=6mm,
  top=4mm,
  bottom=4mm,
  title=#1,
  fonttitle=\bfseries,
  coltitle=black,
  colbacktitle=gray!8,
  attach title to upper,
  before skip=10pt,
  after skip=10pt
}

\section{RoboSteer Benchmark}
\label{Appendix_A}

\subsection{Data Statistics}
\label{sec:data-statistics}
RoboSteer contains 773,716 benchmark instances across 20 task families. Its primary motion collection comprises 38,522 unique human motion sequences, each paired with a source human video and a rendered skeleton video. Order and Times additionally use BABEL-linked AMASS segments, as described in Section~\ref{sec:level2}. The 38,522 human videos span 45.59 hours and 5,359,608 frames, while the 38,522 rendered skeleton videos span 45.99 hours and 1,324,107 frames. These paired video assets cover 91.58 hours and 6,683,715 frames in total. Key-frame Conditioning uses 49,710 extracted images. Each benchmark instance is specified by one task JSON.

The benchmark instances are organized into three levels. Level 1 contains 630,623 instances (81.51\% of the benchmark) across 12 task families: 195,162 full-conditioning reproduction, 203,339 temporal-completion, and 232,122 spatial-completion instances. Level 2 contains 135,329 instances (17.49\%) across seven behavioral constraints. Level 3 contains 7,764 instances (1.00\%) of Interleaved Multi-Source Steering.

\begin{figure}[H]
  \centering
  \includegraphics[width=0.95\textwidth]{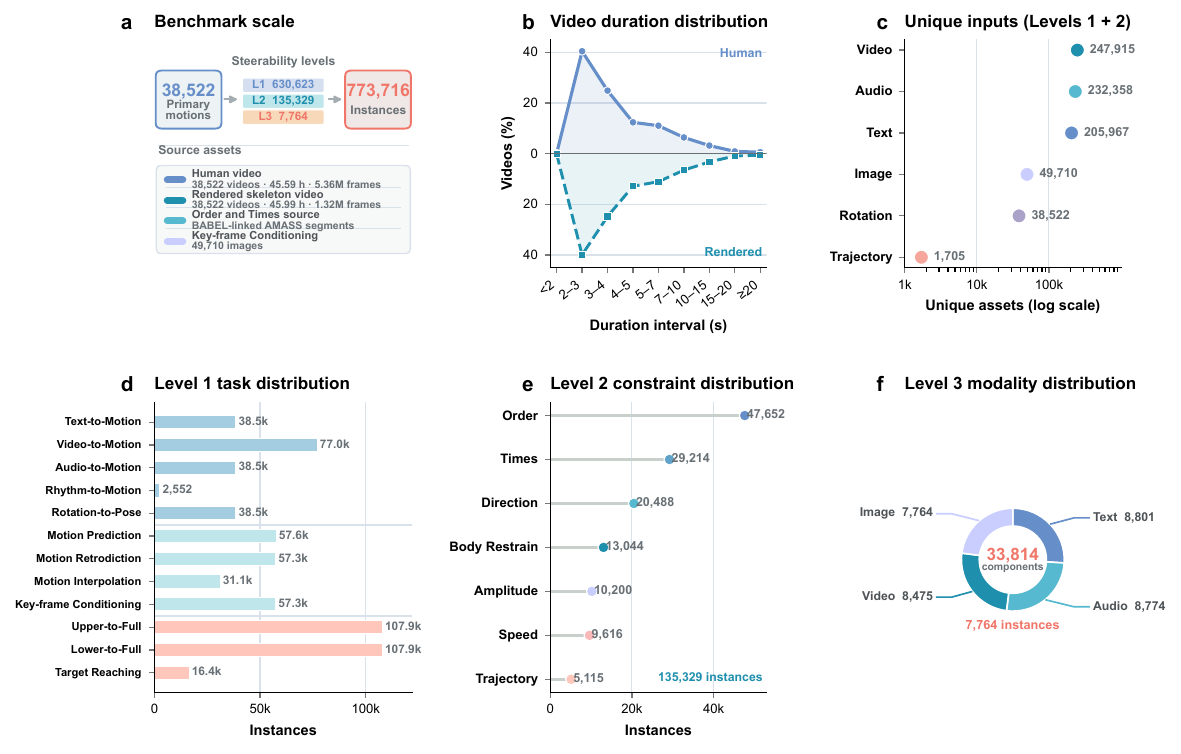}
  \caption{Overview of RoboSteer dataset statistics. \textbf{a}, Benchmark scale and associated media assets across the three steerability levels. \textbf{b}, Duration distributions of source human videos and rendered skeleton videos. \textbf{c}, Unique input assets used by Levels 1 and 2, counted once per source asset within this combined scope. The horizontal axis is logarithmic. \textbf{d}, Distribution of Level 1 task instances. \textbf{e}, Distribution of Level 2 constraint instances. \textbf{f}, Modality components used by Level 3 Interleaved Multi-Source Steering.}
  \label{fig:appendix-statistics}
\end{figure}

Figure~\ref{fig:appendix-statistics} summarizes benchmark scale, video durations, task counts, and modality accounting. Panel a reports instance totals and associated assets. Panel b shows that both source human and rendered skeleton videos are concentrated between 2 and 5 s. Panels d and e report Level 1 task-family and Level 2 constraint counts, respectively. Video-to-Motion Imitation contains 77,044 instances, comprising 38,522 human-video and 38,522 rendered-skeleton-video conditions. Panel f reports the 33,814 ordered modality components in Level 3: 8,801 Text, 8,774 Audio, 8,475 Video, and 7,764 Image components.

Panel c reports unique input assets used by Levels 1 and 2: 247,915 videos, 232,358 audio recordings, 205,967 texts, 49,710 images, 38,522 rotation packages, and 1,705 trajectory images. Repeated references to the same source asset count once. Text entries with different wording remain distinct; fixed task prompts are excluded. Image denotes key frames, Trajectory denotes the Level 2 trajectory images, and Rotation denotes rotation-parameter packages.

\subsection{Dataset Biases}
~\label{sec:data-biases}
The following statistics describe action categories, joint participation, and video annotations in the primary collection of 38,522 source motions.

\textbf{Action-category distribution.} The 38,522 source motion sequences cover 133 fine-grained action categories. Daily, Locomotion, Sports, and Performance account for 18,742 (48.65\%), 5,005 (12.99\%), 4,450 (11.55\%), and 4,131 (10.72\%) sequences, respectively, together comprising 83.92\% of the source motions. Dance, Fitness, Gesture, Martial arts, Gymnastics, and Yoga account for 2.55\%, 2.44\%, 1.49\%, 0.51\%, 0.05\%, and 0.01\%, respectively, while 3,478 sequences (9.03\%) are labeled as \texttt{other}. The distribution is long-tailed: the six most frequent categories---\texttt{daily\_cleaning}, \texttt{daily\_carrying}, \texttt{locomotion\_walking}, \texttt{other}, \texttt{daily\_cooking}, and \texttt{performance\_speech}---jointly account for 65.34\% of the source motions. RoboSteer therefore provides stronger coverage of common daily activities, locomotion, and performance-related motions than of yoga, gymnastics, martial arts, and several fine-grained dance categories.

\textbf{Active-joint and body-region coverage.} A joint degree of freedom is active when its sequence-level angular range of motion is at least 0.30 rad. We summarize each motion type by the median number of active degrees of freedom across its sequences. Of the 133 motion types, 91 (68.42\%) have medians of 20--29, 41 (30.83\%) of 10--19, and one (0.75\%) below 10. Sports, Dance, and Martial arts are dominated by broad multi-joint motion. Gesture and Performance include more motion types with 10--19 active degrees of freedom. Body-region range of motion is averaged across the joints in each region and the sequences in each motion family. The waist has the lowest mean range of motion in every family. Arm motion is particularly pronounced in Dance, Martial arts, Gymnastics, and Yoga, while Locomotion and Fitness show comparatively stronger leg motion.

\textbf{Age and gender annotations.} In the source-video annotations, 76.25\% of subjects are assigned to the 18--39 age range. Male and female labels account for 66.76\% and 33.24\%, corresponding to 25,716 and 12,806 videos, respectively.

\begin{figure}[H]
  \centering
  \includegraphics[width=0.85\textwidth]{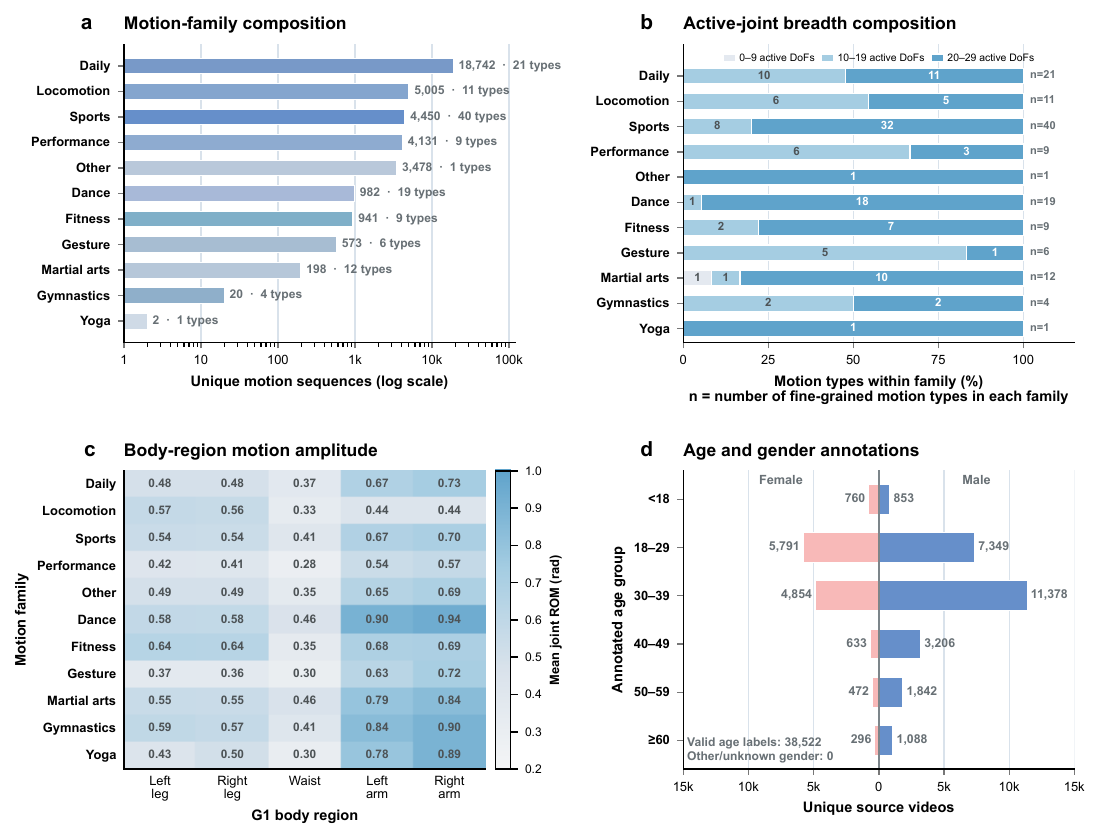}
  \caption{Dataset composition and potential biases. \textbf{a}, Numbers of unique motion sequences and fine-grained motion types in each motion family. \textbf{b}, Composition of motion types by the median number of active joint degrees of freedom within each family (angular range of motion $\geq 0.30$ rad). \textbf{c}, Joint angular range of motion averaged across the joints of each G1 body region and all sequences in each motion family. \textbf{d}, Age and gender distributions from the source-video annotations.}
  \label{fig:appendix-biases}
\end{figure}

Figure~\ref{fig:appendix-biases} summarizes these dataset-level composition and coverage patterns.

\textbf{Task-construction biases.} Task-specific whitelists select motion categories suited to each constraint. Rhythm-to-Motion Alignment retains rhythm-oriented categories and applies an audio-content filter. Direction and Trajectory emphasize motions with root displacement. Speed and Amplitude select actions whose tempo or spatial extent can be modified while preserving action identity. Body Restrain selects actions with a distinguishable secondary body region. Human-video variants of body completion additionally require stable full-body visibility. Section~\ref{sec:task-construction} specifies these filters.

\subsection{Known Limitations}
\label{sec:limitations}
RoboSteer evaluates single-robot motion generation from single-subject motion data. It does not explicitly evaluate physical interaction with people or objects, coordination among multiple robots, or long-horizon planning across successive goals. Its source collection contains more daily activities and locomotion than Yoga and Gymnastics, as quantified in Section~\ref{sec:data-biases}.

Semantic-audio conditions use synthesized speech with the voices and synthesis parameters specified in Section~\ref{sec:caption-audio}. Temporal text and semantic-audio conditions include a global action summary derived from the complete sequence, together with detailed context from the observed intervals. 
\renewcommand{\thetable}{B.\arabic{table}}
\renewcommand{\thefigure}{B.\arabic{figure}}
\setcounter{figure}{0}

\section{Task Construction}
\label{Appendix_B}

\subsection{Data Collection}
\begin{figure}
    \centering
    \includegraphics[width=\linewidth]{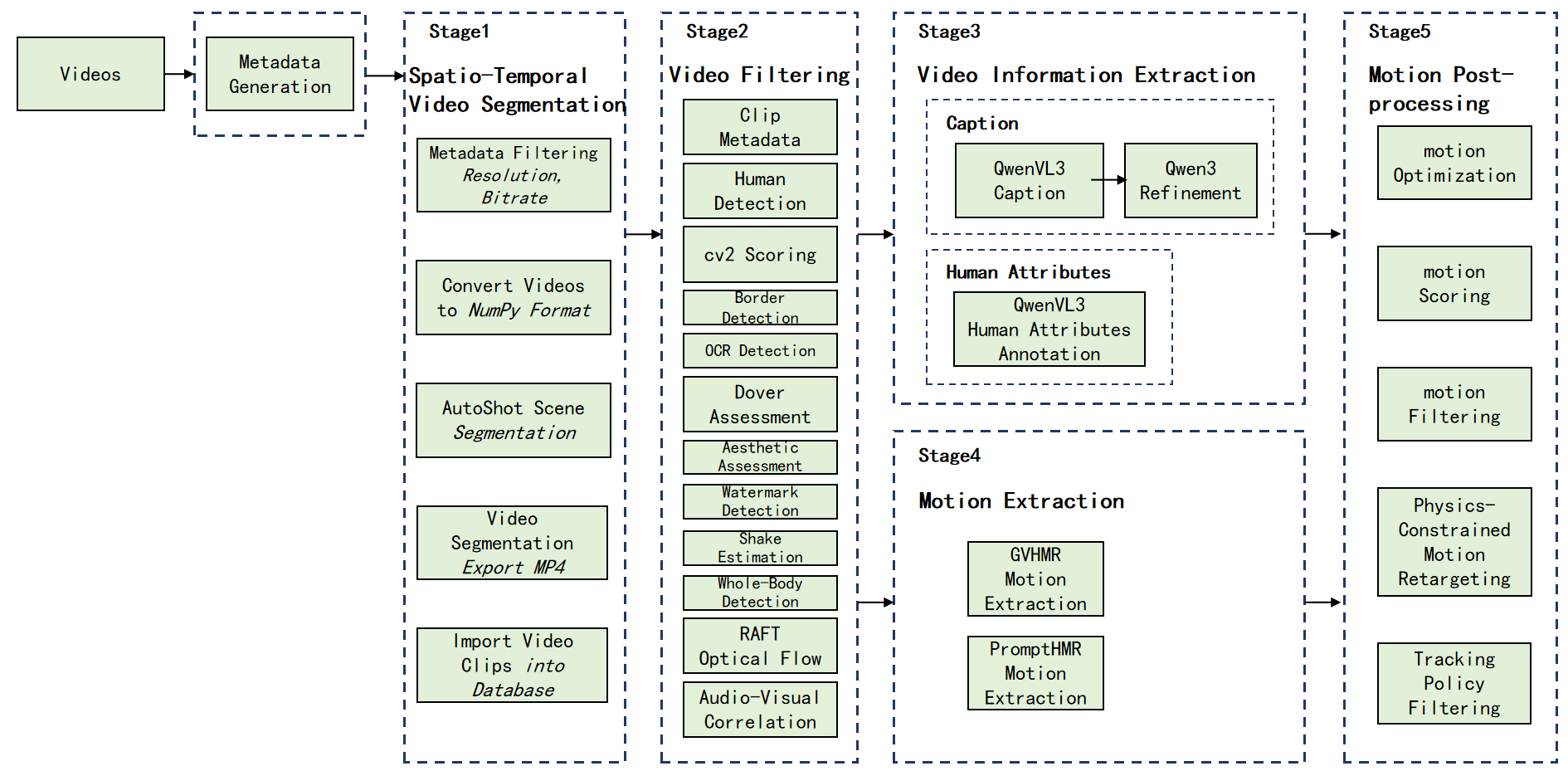}
    \caption{Overview of the raw data collection and processing pipeline. Starting from web-scale-videos, the pipeline performs spatio-temporal segmentation, video filtering, multimodal information extraction, human motion extraction, and motion post-processing to obtain high-quality,
physically executable humanoid motion data.}
    \label{fig:appendix-data-collection}
\end{figure}

We construct the raw motion corpus through a five-stage automated pipeline, as illustrated in Figure ~\ref{fig:appendix-data-collection}. Stage 1 performs spatio-temporal video segmentation and organizes the resulting clips with associated metadata. Stage 2 applies multi-criteria video filtering based on clip metadata, human presence, visual quality, borders, OCR, aesthetics, watermarks, camera motion, whole-body visibility, optical flow, and audio-visual consistency. Stage 3 extracts multimodal information from the retained clips, including refined captions and human attributes. Stage 4 reconstructs human motions from the videos using complementary motion extraction methods. Finally, Stage 5 performs motion optimization, scoring, and filtering, followed by physics-constrained humanoid motion retargeting and tracking-policy-based filtering to obtain executable motions for subsequent benchmark construction.

\subsection{Motion Caption and Semantic-Audio Construction}
\label{sec:caption-audio}

The raw motion descriptions are generated from the original human video associated with each sample. Gemini-3-Flash-Preview processes this video and produces temporally ordered action segments. Its annotation prompt requires temporal boundaries to be placed at clear visual changes or transitions between action states, so that each segment represents a complete, independently describable action. The segments must be chronological, non-overlapping, and collectively cover the entire video without gaps. Their durations adapt to motion rhythm and complexity instead of following fixed temporal windows. The output is constrained to a JSON object containing the action category, start and end times, and body-part-specific descriptions. Each segment contains \texttt{hands}, \texttt{motion}, \texttt{whole body}, \texttt{upper limb}, \texttt{lower limb}, and \texttt{face} descriptions, together with start and end times. We use only \texttt{whole body}, \texttt{upper limb}, and \texttt{lower limb}. The \texttt{hands} and \texttt{face} fields are excluded from subsequent text-task construction. The temporal segmentation supports text selection, video cropping, and ground-truth motion slicing, ensuring aligned action boundaries across modalities.

We select different text fields and temporal intervals for different Level 1 task families. Text-to-Motion Generation and Audio-to-Motion Generation use the \texttt{whole body}, \texttt{upper limb}, and \texttt{lower limb} descriptions from every segment. These descriptions are merged into one chronologically ordered paragraph. Lower-to-Full Body Completion removes upper-limb information and retains only \texttt{whole body} and \texttt{lower limb} descriptions. Upper-to-Full Body Completion removes lower-limb information and retains only \texttt{whole body} and \texttt{upper limb} descriptions.

For temporal completion, let a sample contain $N$ segments and define $k=\lceil N/2\rceil$. When $N\geq2$, Motion Prediction uses the first $k$ segments, whereas Motion Retrodiction uses the last $k$ segments. Motion Interpolation retains only the context at both ends of the sequence. Specifically, when $N\geq3$, we define $s=\lfloor(N-1)/2\rfloor$ and select the first and last $s$ segments. This operation completely excludes one middle segment when $N$ is odd and two middle segments when $N$ is even. The detailed interpolation caption therefore contains no segment descriptions from the hidden middle interval. We additionally use all segments to generate a one-sentence global action summary without body-part details. This summary is paired with the detailed text in Motion Prediction, Motion Retrodiction, and Motion Interpolation. It is also paired with music in Rhythm-to-Motion Alignment and with images in Key-frame Conditioning.

Two models handle complementary caption-polishing operations. GPT-5.5 merges full, prefix, and suffix sequences at a temperature of 0.10. It primarily resolves temporal transitions, removes redundant information, and integrates longer descriptions. Gemini-3-Pro-Preview performs transformations with explicit deletion or formatting constraints, including upper-limb removal, lower-limb removal, global summarization, and interpolation-caption generation. Their respective temperatures are 0.05, 0.05, 0, and 0.05. Low temperatures reduce unsupported changes to motion facts, leakage of excluded body-part information, and output-format variation.

Each generated caption is accepted only after automatic validation. The validator requires a parseable JSON object with a non-empty, non-truncated text value and rejects outputs containing analysis, meta-commentary, or draft markers. Body-completion captions are checked for leakage from the excluded upper- or lower-limb vocabulary. Global summaries are checked for the required one-sentence format. Motion Interpolation captions must contain the exact sentence ``The middle sequence is missing.'' once, with valid context on both sides. A failed output is regenerated for at most three attempts. Samples that remain invalid are recorded in the failure log and are not accepted as valid annotations.

All model outputs use a unified JSON format. Table~\ref{tab:caption-prompts} reports the model, temperature, and complete prompt for each annotation target.

\begingroup
\scriptsize
\begin{longtable}{@{}>{\RaggedRight\arraybackslash}p{0.27\textwidth} >{\RaggedRight\arraybackslash}p{0.17\textwidth} >{\RaggedRight\arraybackslash}p{0.48\textwidth}@{}}
\caption{Models, temperatures, and complete prompts used for motion-caption polishing.}\label{tab:caption-prompts}\\
\toprule
\textbf{Level 1 task family and annotation use} & \textbf{Model and temperature} & \textbf{Complete prompt} \\
\midrule
\endfirsthead
\toprule
\textbf{Level 1 task family and annotation use} & \textbf{Model and temperature} & \textbf{Complete prompt} \\
\midrule
\endhead
\midrule
\multicolumn{3}{r}{\footnotesize Continued on next page}\\
\endfoot
\bottomrule
\endlastfoot
Text-to-Motion Generation and Audio-to-Motion Generation: full-sequence caption; Motion Prediction and Motion Retrodiction: partial-sequence caption & GPT-5.5, $T=0.10$ & \texttt{You will be given motion text from one clip.}\newline \texttt{Your task is to rewrite it into one fluent paragraph.}\newline \texttt{Keep only human motion and essential object interaction.}\newline \texttt{Remove appearance, clothing, face, camera, lighting, and unrelated background details.}\newline \texttt{Do not invent actions or add unsupported details.}\newline \texttt{Be concise and natural.}\newline \texttt{Input: \{raw\_text\}}\newline \texttt{Output only the rewritten paragraph.} \\
\addlinespace[0.25em]
Lower-to-Full Body Completion: remove upper-limb information & Gemini-3-Pro-Preview, $T=0.05$ & \texttt{You will be given motion text from one clip.}\newline \texttt{Your task is to rewrite it into one fluent paragraph.}\newline \texttt{Remove all hand, arm, elbow, wrist, shoulder, finger, thumb, palm, fist, and forearm details.}\newline \texttt{Keep the remaining motion only if it can be expressed without those parts.}\newline \texttt{Do not replace the removed details with vague summary language.}\newline \texttt{Remove appearance, clothing, face, camera, lighting, and unrelated background details.}\newline \texttt{Do not invent actions or add unsupported details.}\newline \texttt{Be concise and natural.}\newline \texttt{Input: \{raw\_text\}}\newline \texttt{Output only the rewritten paragraph.} \\
\addlinespace[0.25em]
Upper-to-Full Body Completion: remove lower-limb information & Gemini-3-Pro-Preview, $T=0.05$ & \texttt{You will be given motion text from one clip.}\newline \texttt{Your task is to rewrite it into one fluent paragraph.}\newline \texttt{Remove all leg, foot, knee, ankle, hip, thigh, calf, and lower-limb details.}\newline \texttt{Keep torso and upper-body motion only if they are present.}\newline \texttt{Do not replace the removed details with vague summary language.}\newline \texttt{Remove appearance, clothing, face, camera, lighting, and unrelated background details.}\newline \texttt{Do not invent actions or add unsupported details.}\newline \texttt{Be concise and natural.}\newline \texttt{Input: \{raw\_text\}}\newline \texttt{Output only the rewritten paragraph.} \\
\addlinespace[0.25em]
Rhythm-to-Motion Alignment and Key-frame Conditioning; Motion Prediction, Motion Retrodiction, and Motion Interpolation: global action summary & Gemini-3-Pro-Preview, $T=0$ & \texttt{You will be given motion text from one clip.}\newline \texttt{Your task is to summarize the main action in exactly one sentence.}\newline \texttt{Start with 'The person is'.}\newline \texttt{Keep the sentence short, direct, and focused on the main motion.}\newline \texttt{If the clip has multiple motions, choose the dominant one instead of listing everything.}\newline \texttt{Do not leave the output blank.}\newline \texttt{Do not mention body parts, appearance, clothing, face, camera, lighting, or unrelated background details.}\newline \texttt{Input: \{raw\_text\}}\newline \texttt{Output only the single sentence.} \\
\addlinespace[0.25em]
Motion Interpolation: interpolation caption & Gemini-3-Pro-Preview, $T=0.05$ & \texttt{You will be given motion text from the beginning and ending parts of the same clip, with the middle part missing.}\newline \texttt{Your task is to rewrite them into one coherent paragraph.}\newline \texttt{Write the first segment first, then insert the exact sentence 'The middle sequence is missing.' once in the middle, then continue with the later segment.}\newline \texttt{Keep only motion facts and essential object interaction.}\newline \texttt{Do not mention appearance, clothing, face, camera, lighting, or unrelated background details.}\newline \texttt{Do not invent actions or add unsupported details.}\newline \texttt{Be concise and natural.}\newline \strut\newline \texttt{Here is one example:}\newline \texttt{Input: The pitcher stands on the mound with his glove held near his chest, then draws his right arm back while stepping forward. He follows through with his arm extended after the throw and settles into a balanced stance.}\newline \texttt{Output: The pitcher stands on the mound with his glove held near his chest, then draws his right arm back while stepping forward. The middle sequence is missing. He follows through with his arm extended after the throw and settles into a balanced stance.}\newline \strut\newline \texttt{Input: \{raw\_text\}}\newline \texttt{Output only the paragraph.} \\
\addlinespace[0.25em]
All annotated Level 1 task families: shared output constraint & --- & \texttt{Return JSON only. Do not include markdown, explanations, analysis, or thought process.}\newline \texttt{Schema: \{"text": "\textless{}the final rewritten caption\textgreater{}"\}}\newline \texttt{The text value must contain only the final caption.} \\
\addlinespace[0.25em]
\end{longtable}
\endgroup

\taskheading{Semantic-Audio Construction}

Semantic audio expresses the motion content or behavioral instruction specified by each task. For the temporal-completion and spatial-completion variants of Level 1, we normalize whitespace in the source condition text and synthesize that text without introducing additional action descriptions. Motion Prediction uses the global action summary and prefix details; Motion Retrodiction uses the summary and suffix details; Motion Interpolation uses the summary and endpoint context, retaining the statement that the middle sequence is missing. Key-frame Conditioning uses the global action and duration information alongside the original images and timestamps. Upper-to-Full and Lower-to-Full Body Completion retain the same body-region omissions as their text variants. Target Reaching includes the base action and local goal.

For Level 2, the spoken condition combines the base-action text with the behavioral modifier. When a separate modifier is present, the two are connected by the sentence ``While performing this action, also follow this instruction:''; otherwise the source text is spoken directly. The audio therefore specifies the requested amplitude, speed, direction, restrained body region, action order, repetition count, or local action associated with a trajectory. The task JSON replaces the source text with an audio-file reference and clears the separate modifier field. Other required assets and ground-truth references are retained, including the images in Key-frame Conditioning and the trajectory image in the Level 2 Trajectory condition.

These Level 1 and Level 2 semantic-audio variants use Edge-TTS with four voices: \path{en-US-AriaNeural}, \path{en-US-GuyNeural}, \path{en-US-JennyNeural}, and \path{en-GB-SoniaNeural}. A SHA-256 digest of a versioned task identifier determines the voice, an integer rate offset from $-15\%$ to $+15\%$, and an integer pitch offset from $-8$ to $+8$ Hz. Audio is stored as MP3. Audio-to-Motion Generation uses the synthesis parameters specified in Section~\ref{sec:level1}. The synthesized-speech duration does not redefine the target motion interval.

\subsection{Task Construction}
\label{sec:task-construction}

Task construction comprises sample selection, input construction, and assignment of motion references or targets. Selection checks the text, temporal segments, visual content, audio content, and motion properties required by each task. Inputs contain the selected text, video, speech, music, image, or rotation parameters. Motion assets retain the source sequence or extract the required temporal interval. Sample identifiers and temporal boundaries link the conditions to their corresponding motion assets.

Retargeted G1 motion packages contain six frame-synchronized CSV files. \texttt{joint\_pos.csv} stores generalized positions for 29 robot-joint degrees of freedom, namely joint angles. \texttt{joint\_vel.csv} stores the corresponding joint velocities. \texttt{body\_pos.csv} records the three-dimensional world coordinates of the root rigid body, whereas \texttt{body\_quat.csv} stores its quaternion orientation. \texttt{body\_lin\_vel.csv} and \texttt{body\_ang\_vel.csv} record root linear and angular velocities, respectively. The accompanying \texttt{info.txt} and \texttt{metadata.txt} files record conversion details, including source and target frame rates, quaternion order, degree-of-freedom order, and total frame count. All six CSV files are converted to 50 Hz and contain the same number of frames. Full-generation tasks use the complete sample folder. Temporal-completion tasks multiply segment times by 50 and round them to frame indices. The same frame interval is then applied to all six files, preserving a complete and synchronized motion-parameter set in each output directory.

\begin{minipage}{\textwidth}
Order and Times use BABEL-linked AMASS motion packs containing frame-wise body poses, root translations, and the source frame rate. Order slices the annotated action--transition--action interval, whereas Times repeats the selected pose and translation sequence in AMASS frame coordinates.
\end{minipage}

\subsubsection{Level 1: Basic Conditional Generation and Local Completion}\label{sec:level1}

Level 1 comprises three task groups: full-conditioning reproduction, temporal completion, and spatial completion with target constraints.

\taskheading{Full-Conditioning Reproduction}

Text-to-Motion Generation. This task uses all 38,522 samples because every sample contains a complete motion caption and its corresponding motion. The input is generated by merging the \texttt{whole body}, \texttt{upper limb}, and \texttt{lower limb} descriptions from all segments through the B.1 polishing procedure. The target asset is the complete motion sequence.

Video-to-Motion Imitation. Both video modalities use all 38,522 samples. The Human variant receives the original human video, whereas the Skeleton variant receives a skeleton video rendered from the same motion sequence. Both variants use the complete motion as the target. No additional semantic filtering is required because every sample directly provides a video-motion pair.

Audio-to-Motion Generation. This task uses all 38,522 samples. We synthesize a spoken condition from the complete motion caption using Edge-TTS. The speaker is sampled from \path{en-US-AriaNeural}, \path{en-US-GuyNeural}, \path{en-US-JennyNeural}, and \path{en-GB-SoniaNeural}. Speech rate is sampled from $-20\%$ to $+25\%$, and pitch is sampled from $-10$ to $+10$ Hz. The synthesized speech is stored as MP3 and used as input; the complete motion is the target.

Rhythm-to-Motion Alignment. This task applies both a motion-type whitelist and an audio-content filter. The whitelist retains dance, periodic sports, fitness, yoga, martial arts, gymnastics, and performance categories. These categories typically exhibit a stable beat, repeated motion cycles, or an explicit rhythmic structure. Table~\ref{tab:rhythm-whitelist} provides the complete whitelist. We then extract the full soundtrack and classify it using the pretrained Audio Spectrogram Transformer (AST) model \texttt{MIT/ast-finetuned-audioset-10-10-0.4593}. The three highest-probability classes are examined in rank order. The first class meeting either threshold determines the decision: a music-related score above 0.3 accepts the sample, and a speech score above 0.5 rejects it. Samples with neither threshold met are rejected. Only samples passing both semantic and audio filters enter the task. The input combines the complete music track with the global action summary, and the target is the complete motion.

Rotation-to-Pose Generation. This task uses all 38,522 samples because complete rotation parameters can be extracted from every retargeted motion. Input assets contain 29-dimensional joint degrees of freedom per frame, the root rotation as a quaternion, and a three-dimensional root-translation offset. All three arrays must have the same frame count. Samples with non-finite values, empty sequences, or all-zero rotations are excluded. The target is the corresponding complete motion sequence.

\taskheading{Temporal Completion}

Motion Prediction requires at least two segments. Let $N$ denote the segment count and define $k=\lceil N/2\rceil$. The text condition combines the global action summary with the polished caption of the first $k$ segments. The video variants crop the corresponding human or skeleton video over the same context interval. The Audio condition synthesizes the same global summary and prefix caption. The boundary is the \texttt{end\_time} of segment $k$. The ground truth is the future interval from this boundary to the end, synchronously sliced from all six 50-Hz CSV files in the corresponding ground-truth motion package.

\begin{minipage}{\textwidth}
Motion Prediction also includes a music-conditioned variant. It uses the motion-type whitelist and AST music detector described in Table~\ref{tab:rhythm-whitelist}, and additionally requires audio longer than 6 s. The input combines the first half of the music with the global action summary. All six 50-Hz CSV files are sliced over the shared interval from the audio midpoint to its end, producing the ground-truth motion for the second half.
\end{minipage}

Motion Retrodiction. This task also requires $N\geq2$ and uses the last $k=\lceil N/2\rceil$ segments as input. The text condition combines the global action summary with the polished caption of these segments. The boundary is the \texttt{start\_time} of the first segment among the last $k$ segments. Human-video and skeleton-video inputs extend from this boundary to the end of the action. The Audio condition synthesizes the same summary and suffix caption as the text condition. The ground truth synchronously retains frames from the action start to the boundary in all six 50-Hz CSV files.

Motion Interpolation. This task requires $N\geq3$. Define $s=\lfloor(N-1)/2\rfloor$. The endpoint context includes only the first and last $s$ segments. The left boundary is the \texttt{end\_time} of segment $s$, and the right boundary is the \texttt{start\_time} of the $s$-th segment from the end. The text condition combines the global action summary with an endpoint caption that inserts the fixed sentence ``The middle sequence is missing.'' between the left and right contexts. The Audio condition synthesizes the same summary and endpoint caption, including the missing-middle statement. Each video variant crops two independent clips outside the two boundaries. The ground truth synchronously extracts all frames between these boundaries from the six 50-Hz CSV files. The detailed caption and both video clips therefore exclude the hidden middle interval, while the global summary provides only action-level background derived from the complete sequence.

Key-frame Conditioning. This task requires at least two segments. We extract a key frame at the end of every segment except the last, capturing visual conditions at action-state transitions. Human and Skeleton variants extract images from the original video and skeleton video, respectively. These images and their timestamps are combined with the global action summary and duration information. Each visual source supports a text-conditioned and an audio-conditioned setting, yielding Human Image+Text, Skeleton Image+Text, Human Image+Audio, and Skeleton Image+Audio. The Audio settings retain the images and timestamps and replace the descriptive text with speech. The complete motion is the target.

\taskheading{Spatial Completion and Target Constraints}

Upper-to-Full Body Completion. The text condition uses all 38,522 samples. Its input retains overall and upper-limb descriptions while removing lower-limb information. The Audio condition synthesizes this same restricted text for all 38,522 cases. Video variants first require stable full-body visibility. We apply YOLOv8n-pose to at most 12 uniformly sampled frames. Upper-body visibility, lower-body visibility, and the valid-pose-frame ratio must each be at least 0.80. For the Human variant, YOLOv8n-pose estimates the left and right hip positions in every frame. Their mean vertical coordinate defines the waist boundary. Original RGB pixels above this boundary are retained, and the region below it is set to black. If valid hips are unavailable in a frame, the most recent valid waist position is reused to prevent abrupt boundary shifts.

The Skeleton variant regenerates and renders a half-body mesh instead of applying a two-dimensional mask to the complete skeleton video. We first pass the frame-wise SMPL-X global orientation, 21 body-joint poses, shape parameters, and root translation to the SMPL-X layer. This recovers the mesh vertices for every frame. Upper-body vertices are then selected using the SMPL-X linear-blend-skinning weights. We retain vertices whose weights exceed 0.1 for the pelvis, three spine joints, neck, head, arms, or hands. A triangle is retained only when all three of its vertices belong to this set. This definition preserves the full torso and upper limbs from the pelvic waist boundary while removing the leg mesh. Finally, \texttt{smplx\_upper.faces} and the complete vertex sequence are rendered in Blender to produce the upper-body video.

Lower-to-Full Body Completion. The text condition uses all 38,522 samples. Its input retains overall and lower-limb descriptions while removing upper-limb information. The Audio condition synthesizes this same restricted text for all 38,522 cases. The Human variant uses the same visibility filter and frame-wise waist localization. It retains RGB pixels below the waist boundary and sets the region above it to black. The Skeleton variant retains vertices whose skinning weights exceed 0.1 for the pelvis, hips, knees, ankles, or feet, while explicitly excluding spine joints. Triangles are retained only when all three vertices belong to this set. \texttt{smplx\_lower.faces} and the same vertex sequence are then rendered to produce a video containing only the mesh below the waist. Both spatial-completion tasks use the complete, unmasked motion as ground truth.

Target Reaching. This task selects skeleton videos with an explicit internal body-part goal. Gemini-3-Flash-Preview first identifies a clear and intentional relation between body parts, such as a hand touching the chest. Samples showing only proximity, incidental passage, or uncertain contact are rejected. The same model then independently verifies that the specified target is clearly visible without selecting an alternative target. Finally, the model determines whether the local target can be separated from the primary action. Samples are rejected if the target is necessary to complete the primary action. Otherwise, direct and synonymous descriptions of the target are removed from the base-action caption. Every model stage uses Gemini-3-Flash-Preview with temperature 0 and requires JSON output. The text condition presents separate \texttt{Base Action} and \texttt{Local Goal} descriptions. The Audio condition expresses the same base action and local goal in speech. Each setting contains 8,193 instances and uses the original complete motion as the target.

\subsubsection{Level 2: Controllable Motion Modification and Composition}\label{sec:level2}

Amplitude, Speed, Direction, and Body Restrain provide text, human-video, rendered-skeleton-video, and audio conditions for a base action and a behavioral constraint. Their \texttt{ground\_truth} fields reference the unmodified source motion used by Level 1. Constraint following is assessed by comparing properties of the generated motion with this reference; for Speed, the comparison uses motion duration. Order and Times use targets constructed from temporally annotated BABEL--AMASS segments.

\taskheading{Speed}

We first use motion type to retain actions that permit a clear speed transformation without changing their identity. Eligible actions have observable cadence, motion frequency, or repeated rhythm. We exclude actions that depend on extreme dynamics, fixed equipment, complex environments, or rigid stylistic forms. Table~\ref{tab:level2-whitelists} provides the whitelist. For each sample, we recover trajectories for 22 three-dimensional SMPL-X joints. Absolute and root-relative joint speeds are computed from successive three-dimensional joint displacements multiplied by the frame rate, in metres per second. Each quantity is averaged over frames and the core joints of the motion type. The speed score is the mean of these two quantities:

\begin{equation*}
S_{\mathrm{speed}} = \frac{\overline{v_{\mathrm{abs}}}+\overline{v_{\mathrm{rel}}}}{2}
\end{equation*}

Scores are converted to percentiles within each motion type. We retain only samples in the 20th--40th and 60th--80th percentile bands. The former receive the instruction to perform the action at twice its original speed. The latter receive the instruction to perform it at half its original speed. This selection avoids samples already near the extremes of their motion-type speed distribution. Each retained sample yields text, human-video, rendered-skeleton-video, and audio variants. The Audio condition speaks the base action and the requested speed factor.

\taskheading{Direction}

We first use motion type to retain actions that can generally produce sustained horizontal displacement. These include core locomotion, slope or stair locomotion, and extended forms such as skateboarding, roller skating, cycling, skiing, and rowing. Table~\ref{tab:level2-whitelists} lists the complete whitelist and semantic groups. We then geometrically filter the root trajectory in the horizontal plane. Path length must be at least 0.10 m, and endpoint displacement must be at least 0.20 m. The displacement-to-path-length ratio must be at least 0.80, and PCA linearity must be at least 0.95. The normalized line-fitting residual must not exceed 0.05, and absolute cumulative turning must not exceed 0.75 rad.

\begin{minipage}{\textwidth}
We estimate the body-facing direction from the root rotations in the first 10 frames and require a facing-consistency score of at least 0.80. The angle between net displacement and initial facing must not exceed $30^\circ$. Net forward displacement must be at least 0.20 m, and its ratio to path length must be at least 0.70. The proportion of backward motion must not exceed 0.10. Samples lacking root rotations or exhibiting an unstable initial facing direction are excluded.
\end{minipage}

Each retained motion produces left and right constraints. These constraints redirect the original forward motion to the left or right while preserving its action content. The task has text, human-video, rendered-skeleton-video, and audio variants. The Audio condition speaks the base action and left or right constraint. Constraint labels require no language-model annotation.

\taskheading{Body Restrain}

Body Restrain uses both motion-type semantics and the observed motion sequence. We first infer the primary and secondary body regions needed to perform each motion type. Locomotion, for example, typically uses the legs as its primary region and the arms as a secondary region. Gesture categories usually reverse this assignment. We retain only categories with a clear non-primary region whose restraint does not obscure the identity of the main action. Table~\ref{tab:level2-whitelists} provides the whitelist. A fixed 22-joint SMPL-X mapping converts body regions into joint indices:

\begingroup
\footnotesize
\begin{center}
\begin{tabular}{@{}>{\RaggedRight\arraybackslash}p{0.17\textwidth} >{\RaggedRight\arraybackslash}p{0.24\textwidth} >{\RaggedRight\arraybackslash}p{0.51\textwidth}@{}}
\toprule
\textbf{Body region} & \textbf{SMPL-X joint indices} & \textbf{Corresponding joints} \\
\midrule
\texttt{pelvis} & 0 & pelvis \\
\addlinespace[0.25em]
\texttt{torso} & 0, 3, 6, 9, 12, 15 & pelvis, spine1, spine2, spine3, neck, head \\
\addlinespace[0.25em]
\texttt{head} & 12, 15 & neck, head \\
\addlinespace[0.25em]
\texttt{shoulders} & 13, 14, 16, 17 & left/right collar and left/right shoulder \\
\addlinespace[0.25em]
\texttt{arms} & 13, 14, 16, 17, 18, 19, 20, 21 & bilateral collars, shoulders, elbows, and wrists \\
\addlinespace[0.25em]
\texttt{hands} & 20, 21 & left/right wrist \\
\addlinespace[0.25em]
\texttt{legs} & 1, 2, 4, 5, 7, 8, 10, 11 & bilateral hips, knees, ankles, and feet \\
\addlinespace[0.25em]
\texttt{feet} & 7, 8, 10, 11 & bilateral ankles and feet \\
\addlinespace[0.25em]
\bottomrule
\end{tabular}
\end{center}
\endgroup

For each motion type, the script first expands \texttt{secondary\_groups} through this mapping to obtain \texttt{secondary\_joint\_ids}. These joints are used to calculate the activity score of secondary body regions. The task then selects the final restraint target according to the motion family. Locomotion primarily restrains the arms. Gestures, daily activities, and performances dominated by the torso or upper body primarily restrain the legs. Dance, fitness, martial arts, and gymnastics restrain the arms or legs according to their primary body region. Final prompts allow only \texttt{arms} and \texttt{legs}. Therefore, \texttt{hands} is expanded to \texttt{arms}, and \texttt{feet} is expanded to \texttt{legs}. Table~\ref{tab:body-restrain-targets} lists the final target for every motion type.

We compute absolute speed, spatial range of motion, and root-relative speed over the secondary joints. Spatial range of motion is the Euclidean norm of each joint's coordinate-wise position ranges, measured in metres. The score combines the numerical values of these three terms and is used for within-type ranking:

\begin{equation*}
S_{\mathrm{restrain}} = -\frac{\overline{v_{\mathrm{abs}}^{\mathrm{secondary}}}+\overline{\mathrm{ROM}^{\mathrm{secondary}}}+\overline{v_{\mathrm{rel}}^{\mathrm{secondary}}}}{3}
\end{equation*}

The negative sign reverses the activity ordering, so weaker secondary-region motion receives a higher raw score. Scores are converted to percentiles within each motion type, and only the 35th--65th percentile band is retained. This interval selects actions with moderate activity in the secondary region. If that region is already nearly stationary, adding a restraint has little practical effect. If its activity is too strong, the region may contribute to the main action and be unsuitable for restraint.

\begin{minipage}{\textwidth}
After filtering, the task asks the model to preserve the primary action while keeping the arms or legs specified in Table~\ref{tab:body-restrain-targets} as still as possible. We construct text, human-video, rendered-skeleton-video, and audio variants. The Audio condition names the base action and the body region to keep still.
\end{minipage}

\taskheading{Amplitude}

Amplitude uses motion type to retain actions that support clear spatial scaling through stride length, limb extension, or postural expansion. We exclude actions already near their motion limits, actions dependent on complex objects or environments, static or subtle motions, and rigidly standardized forms. Table~\ref{tab:level2-whitelists} provides the whitelist. For each sample, we use the mean absolute speed, spatial range of motion, and root-relative speed defined above, averaged over the core joints. Their numerical values form the amplitude ranking score:

\begin{equation*}
S_{\mathrm{amplitude}} = \frac{\overline{v_{\mathrm{abs}}}+\overline{\mathrm{ROM}}+\overline{v_{\mathrm{rel}}}}{3}
\end{equation*}

Within each motion type, we retain only the 20th--40th and 60th--80th percentile bands. Low-amplitude samples receive the instruction to perform the action with twice the original amplitude. High-amplitude samples receive the instruction to use half the original amplitude. The task produces text, human-video, rendered-skeleton-video, and audio variants. The Audio condition speaks the base action and the requested amplitude factor.

\taskheading{Trajectory}

Trajectory uses motion type to retain actions that may produce reusable root paths. These include ground locomotion, wheeled or board sports, gliding or riding, and selected ball sports, fitness, martial arts, gymnastics, and daily activities. Table~\ref{tab:level2-whitelists} lists the semantic groups and complete whitelist. After semantic filtering, the horizontal root path must be at least 0.8 m long. We normalize the trajectory by the initial root orientation, aligning the initial facing direction with a shared forward axis. The path is then resampled by arc length, smoothed, and simplified using small-turn suppression to reduce local noise.

We render the processed two-dimensional trajectory and classify its quality using Gemini-3-Flash-Preview at temperature 0. The prompt permits only three clear categories: S-shaped, back-and-forth, and clockwise-or-counterclockwise. Approximately straight, weakly curved, disordered, repeatedly overlapping, or ambiguous trajectories are labeled as low quality. S-shaped trajectories undergo an additional slope-direction test over four equal-arc-length intervals. Each interval must contain at least three valid slopes, and at least 0.90 of them must share the dominant sign.

For the text variant, Gemini-3-Flash-Preview at temperature 0 removes global path shape, directional displacement, and start or endpoint positions from the \texttt{whole body} segments. It retains only local actions, poses, gait, and object interactions. The final input combines the trajectory-free base-action caption with a trajectory image, or uses a processed trajectory video. The Image+Audio setting retains the trajectory image and synthesizes the trajectory-free base-action caption together with any accompanying instruction. Target assets contain the complete motion and normalized trajectory points at arc-length fractions $0$, $1/4$, $1/2$, $3/4$, and $1$.

\taskheading{Order}

Order is constructed from frame-level BABEL action annotations. We identify contiguous ``action A, transition, action B'' triplets in which A and B are not transitions. The intervening transition must last 0.15--1.65 s. Samples are retained only when A and B have different action labels and both the AMASS motion and original video are available.

Each sample receives two semantically equivalent text inputs: \texttt{A then B} and \texttt{do B after doing A}. The original video is cropped from the start of action A to the end of action B. AMASS poses and root translations are sliced over the same interval, and the initial horizontal root position is set to zero. The Audio setting synthesizes the action-order instruction. The Video setting supplies two separate input clips for the ordered actions and retains the ordering prompt. The first clip extends from action A to the end of the transition, and the second starts at the transition and extends to the end of action B; the transition interval is shared. Text, Audio, and Video each contain 15,884 instances. The labels require no language-model annotation.

\taskheading{Times}

Times is constructed from single-action BABEL segments. We first exclude transitions, missing action labels, segments shorter than 0.6 s or longer than 5.0 s, and samples without corresponding AMASS motion. Both endpoints must be close to a stable standing pose. Excessive changes in root height, orientation, or body pose between the endpoints are disallowed. Action-label rules then remove static poses, non-countable actions, and segments that already contain multiple repetitions.

During final filtering, root-orientation difference must not exceed 0.25 rad, and root-height difference must not exceed 0.08 m. Mean standing-joint pose difference must not exceed 0.30, and full-body pose difference must not exceed 0.45. Each retained single action is repeated two, three, and four times. Pose sequences are concatenated directly, whereas horizontal root translation accumulates continuously across repetitions. This construction prevents the root from returning to the origin at the start of each repetition.

Text inputs follow \texttt{repeat \{action\} \{count\} times}, whereas video inputs follow \texttt{repeat the action shown in the video \{count\} times}. The reference video shows the original single-action clip, and the target motion is the concatenated repeated sequence. The Audio condition synthesizes the action label and repetition count from the text condition, preserving the repeated-motion target. Text, Video, and Audio each contain 9,738 instances. The task uses deterministic labels and motion-based filters without language-model annotation.

\taskheading{Motion-type Semantic Whitelists}

Table~\ref{tab:rhythm-whitelist} gives the complete motion-type whitelist shared by Rhythm-to-Motion Alignment and the music-conditioned Motion Prediction setting. It first identifies categories that may exhibit a musical beat, repeated motion cycles, or an explicit performance rhythm. Samples must still pass AST-based audio detection. Motion type therefore defines only the candidate pool and does not replace validation of the original soundtrack.

\begingroup
\scriptsize
\begin{longtable}{@{}>{\RaggedRight\arraybackslash}p{0.14\textwidth} >{\RaggedRight\arraybackslash}p{0.27\textwidth} >{\RaggedRight\arraybackslash}p{0.51\textwidth}@{}}
\caption{Motion-type whitelist shared by Rhythm-to-Motion Alignment and its prediction setting.}\label{tab:rhythm-whitelist}\\
\toprule
\textbf{Category} & \textbf{Rationale} & \textbf{Motion-type whitelist} \\
\midrule
\endfirsthead
\toprule
\textbf{Category} & \textbf{Rationale} & \textbf{Motion-type whitelist} \\
\midrule
\endhead
\midrule
\multicolumn{3}{r}{\footnotesize Continued on next page}\\
\endfoot
\bottomrule
\endlastfoot
Dance & Dance actions are commonly driven by a stable musical beat and contain continuous or repeated motion phrases. & \texttt{dance\_hip\_hop}, \texttt{dance\_street}, \texttt{dance\_breaking}, \texttt{dance\_popping}, \texttt{dance\_locking}, \texttt{dance\_house}, \texttt{dance\_waacking}, \texttt{dance\_voguing}, \texttt{dance\_kpop}, \texttt{dance\_ballet}, \texttt{dance\_contemporary}, \texttt{dance\_modern}, \texttt{dance\_jazz}, \texttt{dance\_tap}, \texttt{dance\_latin}, \texttt{dance\_salsa}, \texttt{dance\_bachata}, \texttt{dance\_cha\_cha}, \texttt{dance\_ballroom}, \texttt{dance\_waltz}, \texttt{dance\_tango}, \texttt{dance\_foxtrot}, \texttt{dance\_traditional}, \texttt{dance\_folk}, \texttt{dance\_ethnic}, \texttt{dance\_social}, \texttt{dance\_party}, \texttt{dance\_freestyle} \\
\addlinespace[0.25em]
Sports & These sports exhibit periodic cadence, striking rhythm, gliding cycles, or continuous coordination patterns. & \texttt{sports\_basketball}, \texttt{sports\_football}, \texttt{sports\_volleyball}, \texttt{sports\_tennis}, \texttt{sports\_badminton}, \texttt{sports\_table\_tennis}, \texttt{sports\_golf}, \texttt{sports\_baseball}, \texttt{sports\_softball}, \texttt{sports\_bowling}, \texttt{sports\_billiards}, \texttt{sports\_skateboarding}, \texttt{sports\_longboarding}, \texttt{sports\_roller\_skating}, \texttt{sports\_inline\_skating}, \texttt{sports\_bmx}, \texttt{sports\_scooter}, \texttt{sports\_parkour}, \texttt{sports\_surfing}, \texttt{sports\_windsurfing}, \texttt{sports\_kitesurfing}, \texttt{sports\_snowboarding}, \texttt{sports\_skiing}, \texttt{sports\_ice\_skating}, \texttt{sports\_speed\_skating}, \texttt{sports\_swimming}, \texttt{sports\_diving}, \texttt{sports\_rowing}, \texttt{sports\_kayaking}, \texttt{sports\_canoeing}, \texttt{sports\_paddleboarding}, \texttt{sports\_cycling}, \texttt{sports\_mountain\_biking}, \texttt{sports\_horse\_riding}, \texttt{sports\_archery}, \texttt{sports\_climbing}, \texttt{sports\_bouldering}, \texttt{sports\_jumping}, \texttt{sports\_throwing}, \texttt{sports\_shooting}, \texttt{sports\_fishing} \\
\addlinespace[0.25em]
Fitness and yoga & Fitness and yoga actions are often organized by repetition count, breathing rhythm, or fixed motion sequences. & \texttt{fitness\_gym}, \texttt{fitness\_aerobics}, \texttt{fitness\_hiit}, \texttt{fitness\_crossfit}, \texttt{fitness\_pilates}, \texttt{fitness\_bodyweight}, \texttt{fitness\_weightlifting}, \texttt{fitness\_powerlifting}, \texttt{fitness\_jump\_training}, \texttt{fitness\_jumping\_rope}, \texttt{fitness\_resistance\_band}, \texttt{fitness\_rehabilitation}, \texttt{yoga\_hatha}, \texttt{yoga\_vinyasa}, \texttt{yoga\_power}, \texttt{yoga\_ashtanga}, \texttt{yoga\_yin} \\
\addlinespace[0.25em]
Martial arts & Martial arts commonly contain explicit offensive and defensive rhythms, routines, or repeated practice cycles. & \texttt{martial\_tai\_chi}, \texttt{martial\_qigong}, \texttt{martial\_kungfu}, \texttt{martial\_taekwondo}, \texttt{martial\_karate}, \texttt{martial\_judo}, \texttt{martial\_boxing}, \texttt{martial\_kickboxing}, \texttt{martial\_mma}, \texttt{martial\_muay\_thai}, \texttt{martial\_wrestling}, \texttt{martial\_fencing}, \texttt{martial\_staff}, \texttt{martial\_sword} \\
\addlinespace[0.25em]
Gymnastics and performance & Gymnastics, conducting, acrobatics, and cheerleading exhibit explicit performance timing and motion rhythm. & \texttt{gymnastics\_floor}, \texttt{gymnastics\_apparatus}, \texttt{gymnastics\_rhythmic}, \texttt{gymnastics\_trampoline}, \texttt{performance\_conducting}, \texttt{performance\_acrobatics}, \texttt{performance\_cheerleading} \\
\addlinespace[0.25em]
\end{longtable}
\endgroup

Table~\ref{tab:level2-whitelists} gives the motion-type whitelists used by the Level 2 tasks. The lists follow the semantic groups defined in the construction scripts. Samples must still pass the subsequent motion-statistical or geometric filters before entering the final tasks.

\begingroup
\scriptsize
\begin{longtable}{@{}>{\RaggedRight\arraybackslash}p{0.20\textwidth} >{\RaggedRight\arraybackslash}p{0.25\textwidth} >{\RaggedRight\arraybackslash}p{0.47\textwidth}@{}}
\caption{Motion-type whitelists used by the Level 2 task families.}\label{tab:level2-whitelists}\\
\toprule
\textbf{Task and semantic group} & \textbf{Rationale} & \textbf{Motion-type whitelist} \\
\midrule
\endfirsthead
\toprule
\textbf{Task and semantic group} & \textbf{Rationale} & \textbf{Motion-type whitelist} \\
\midrule
\endhead
\midrule
\multicolumn{3}{r}{\footnotesize Continued on next page}\\
\endfoot
\bottomrule
\endlastfoot
Speed & These actions have an adjustable cadence, frequency, or repeated rhythm whose modification preserves action identity. & \texttt{locomotion\_walking}, \texttt{locomotion\_brisk\_walking}, \texttt{locomotion\_jogging}, \texttt{locomotion\_hiking}, \texttt{locomotion\_stairs}, \texttt{locomotion\_treadmill}, \texttt{locomotion\_uphill}, \texttt{locomotion\_downhill}, \texttt{gesture\_waving}, \texttt{gesture\_clapping}, \texttt{fitness\_gym}, \texttt{fitness\_aerobics}, \texttt{fitness\_bodyweight}, \texttt{fitness\_crossfit}, \texttt{martial\_kungfu}, \texttt{martial\_boxing}, \texttt{martial\_karate}, \texttt{martial\_taekwondo}, \texttt{dance\_freestyle}, \texttt{dance\_party} \\
\addlinespace[0.25em]
Amplitude & These actions have interpretable stride, extension, or postural range without depending on fixed objects, extreme dynamics, or rigid environmental constraints. & \texttt{locomotion\_walking}, \texttt{locomotion\_brisk\_walking}, \texttt{locomotion\_jogging}, \texttt{locomotion\_hiking}, \texttt{gesture\_communication}, \texttt{gesture\_greeting}, \texttt{gesture\_pointing}, \texttt{gesture\_waving}, \texttt{fitness\_aerobics}, \texttt{fitness\_bodyweight}, \texttt{fitness\_pilates}, \texttt{fitness\_rehabilitation}, \texttt{martial\_tai\_chi}, \texttt{martial\_qigong}, \texttt{martial\_kungfu}, \texttt{martial\_karate}, \texttt{martial\_taekwondo}, \texttt{performance\_conducting}, \texttt{dance\_freestyle}, \texttt{dance\_party}, \texttt{dance\_social}, \texttt{daily\_stretching} \\
\addlinespace[0.25em]
Body Restrain & These actions contain secondary arm or leg motion distinguishable from the primary action, which remains identifiable after restraint. & \texttt{locomotion\_walking}, \texttt{locomotion\_brisk\_walking}, \texttt{locomotion\_jogging}, \texttt{locomotion\_hiking}, \texttt{gesture\_communication}, \texttt{gesture\_greeting}, \texttt{gesture\_pointing}, \texttt{gesture\_waving}, \texttt{fitness\_aerobics}, \texttt{fitness\_bodyweight}, \texttt{fitness\_pilates}, \texttt{fitness\_rehabilitation}, \texttt{martial\_tai\_chi}, \texttt{martial\_qigong}, \texttt{martial\_kungfu}, \texttt{martial\_karate}, \texttt{martial\_taekwondo}, \texttt{performance\_conducting}, \texttt{dance\_freestyle}, \texttt{dance\_party}, \texttt{dance\_social}, \texttt{daily\_stretching}, \texttt{dance\_ballet}, \texttt{dance\_breaking}, \texttt{dance\_popping}, \texttt{dance\_locking}, \texttt{dance\_waacking}, \texttt{dance\_voguing}, \texttt{dance\_tap}, \texttt{gymnastics\_floor}, \texttt{gymnastics\_rhythmic}, \texttt{performance\_cheerleading}, \texttt{performance\_acrobatics}, \texttt{performance\_mime}, \texttt{performance\_magic}, \texttt{performance\_modeling} \\
\addlinespace[0.25em]
Direction: core locomotion & The action semantics directly imply sustained ground displacement. & \texttt{locomotion\_walking}, \texttt{locomotion\_brisk\_walking}, \texttt{locomotion\_jogging}, \texttt{locomotion\_running}, \texttt{locomotion\_sprinting}, \texttt{locomotion\_hiking}, \texttt{locomotion\_crawling}, \texttt{locomotion\_shuffling}, \texttt{locomotion\_limping} \\
\addlinespace[0.25em]
Direction: extended locomotion & Slopes, stairs, wheels, gliding, or rowing generate measurable horizontal displacement. & \texttt{locomotion\_uphill}, \texttt{locomotion\_downhill}, \texttt{locomotion\_stairs}, \texttt{locomotion\_treadmill}, \texttt{sports\_skateboarding}, \texttt{sports\_longboarding}, \texttt{sports\_roller\_skating}, \texttt{sports\_inline\_skating}, \texttt{sports\_bmx}, \texttt{sports\_scooter}, \texttt{sports\_cycling}, \texttt{sports\_mountain\_biking}, \texttt{sports\_skiing}, \texttt{sports\_snowboarding}, \texttt{sports\_ice\_skating}, \texttt{sports\_speed\_skating}, \texttt{sports\_surfing}, \texttt{sports\_windsurfing}, \texttt{sports\_kitesurfing}, \texttt{sports\_rowing}, \texttt{sports\_kayaking}, \texttt{sports\_canoeing}, \texttt{sports\_paddleboarding} \\
\addlinespace[0.25em]
Trajectory: ground locomotion & Ground locomotion can produce straight, curved, back-and-forth, or composite root paths. & \texttt{locomotion\_walking}, \texttt{locomotion\_brisk\_walking}, \texttt{locomotion\_jogging}, \texttt{locomotion\_running}, \texttt{locomotion\_sprinting}, \texttt{locomotion\_hiking}, \texttt{locomotion\_uphill}, \texttt{locomotion\_downhill}, \texttt{locomotion\_stairs}, \texttt{locomotion\_treadmill}, \texttt{locomotion\_crawling}, \texttt{locomotion\_shuffling}, \texttt{locomotion\_limping} \\
\addlinespace[0.25em]
Trajectory: wheeled or board & Wheeled and board actions typically produce continuous and recognizable planar paths. & \texttt{sports\_skateboarding}, \texttt{sports\_longboarding}, \texttt{sports\_roller\_skating}, \texttt{sports\_inline\_skating}, \texttt{sports\_ice\_skating}, \texttt{sports\_speed\_skating}, \texttt{sports\_bmx}, \texttt{sports\_scooter}, \texttt{sports\_cycling}, \texttt{sports\_mountain\_biking} \\
\addlinespace[0.25em]
Trajectory: glide or equine & Gliding, paddling, swimming, and riding may produce long curved or back-and-forth paths. & \texttt{sports\_snowboarding}, \texttt{sports\_skiing}, \texttt{sports\_surfing}, \texttt{sports\_windsurfing}, \texttt{sports\_kitesurfing}, \texttt{sports\_rowing}, \texttt{sports\_kayaking}, \texttt{sports\_canoeing}, \texttt{sports\_paddleboarding}, \texttt{sports\_swimming}, \texttt{sports\_diving}, \texttt{sports\_horse\_riding} \\
\addlinespace[0.25em]
Trajectory: field or court sports & Field and court sports may contain turns, curves, and short-distance displacement. & \texttt{sports\_basketball}, \texttt{sports\_football}, \texttt{sports\_volleyball}, \texttt{sports\_tennis}, \texttt{sports\_badminton}, \texttt{sports\_table\_tennis}, \texttt{sports\_softball}, \texttt{sports\_parkour} \\
\addlinespace[0.25em]
Trajectory: fitness travel or jump & Selected training actions contain usable ground displacement or jumping paths. & \texttt{fitness\_aerobics}, \texttt{fitness\_hiit}, \texttt{fitness\_crossfit}, \texttt{fitness\_jump\_training}, \texttt{fitness\_jumping\_rope} \\
\addlinespace[0.25em]
Trajectory: martial arts or gymnastics & Martial arts, gymnastics, and performance motions may form curved, reversing, or closed root paths. & \texttt{martial\_kungfu}, \texttt{martial\_taekwondo}, \texttt{martial\_karate}, \texttt{martial\_boxing}, \texttt{martial\_kickboxing}, \texttt{martial\_mma}, \texttt{martial\_muay\_thai}, \texttt{martial\_fencing}, \texttt{martial\_staff}, \texttt{martial\_sword}, \texttt{gymnastics\_floor}, \texttt{gymnastics\_apparatus}, \texttt{gymnastics\_rhythmic}, \texttt{gymnastics\_trampoline}, \texttt{performance\_acrobatics}, \texttt{performance\_cheerleading}, \texttt{performance\_modeling} \\
\addlinespace[0.25em]
Trajectory: daily travel or object motion & These daily actions may include body translation or paths caused by carrying, pushing, or pulling objects. & \texttt{daily\_opening\_door}, \texttt{daily\_carrying}, \texttt{daily\_lifting}, \texttt{daily\_pushing}, \texttt{daily\_pulling}, \texttt{daily\_cleaning}, \texttt{daily\_washing} \\
\addlinespace[0.25em]
\end{longtable}
\endgroup

Table~\ref{tab:body-restrain-targets} gives the final task constraint for every motion type in the Body Restrain whitelist. Starting from \texttt{secondary\_groups}, the mapping expands hand constraints to the arms and foot constraints to the legs. This avoids overly small end-effector targets that are difficult to evaluate reliably.

\begingroup
\scriptsize
\refstepcounter{table}\label{tab:body-restrain-targets}
\begin{center}
\textbf{Table \thetable:} Final restraint target for each motion type in the Body Restrain whitelist.

\vspace{0.35em}
\begin{tabular}{@{}>{\RaggedRight\arraybackslash}p{0.46\textwidth} >{\RaggedRight\arraybackslash}p{0.46\textwidth}@{}}
\toprule
\textbf{Arms} & \textbf{Legs} \\
\midrule
\texttt{dance\_breaking}, \texttt{dance\_ballet}, \texttt{dance\_tap}, \texttt{dance\_social}, \texttt{dance\_party}, \texttt{dance\_freestyle}, \texttt{fitness\_aerobics}, \texttt{fitness\_pilates}, \texttt{fitness\_bodyweight}, \texttt{fitness\_rehabilitation}, \texttt{martial\_tai\_chi}, \texttt{martial\_qigong}, \texttt{martial\_kungfu}, \texttt{locomotion\_walking}, \texttt{locomotion\_brisk\_walking}, \texttt{locomotion\_jogging}, \texttt{locomotion\_hiking}, \texttt{gymnastics\_floor}, \texttt{gymnastics\_rhythmic}
&
\texttt{dance\_popping}, \texttt{dance\_locking}, \texttt{dance\_waacking}, \texttt{dance\_voguing}, \texttt{martial\_taekwondo}, \texttt{martial\_karate}, \texttt{performance\_conducting}, \texttt{performance\_magic}, \texttt{performance\_mime}, \texttt{performance\_modeling}, \texttt{performance\_acrobatics}, \texttt{performance\_cheerleading}, \texttt{gesture\_communication}, \texttt{gesture\_greeting}, \texttt{gesture\_pointing}, \texttt{gesture\_waving}, \texttt{daily\_stretching} \\
\bottomrule
\end{tabular}
\end{center}
\endgroup

\subsubsection{Level 3: Interleaved Multi-Source Steering}\label{sec:level3}

Interleaved Multi-Source Steering combines temporally ordered Text, Audio, Video, and Image conditions into a single task whose target is one continuous motion sequence. Candidate samples require an available source video and motion package, a positive duration, and at least three valid caption segments with non-empty whole-body descriptions and positive temporal intervals.

A deterministic seed derived from the sample identifier and planning version assigns one of Text, Audio, or Video to each segment. Every task includes at least one segment of each of these three types. The final segment is assigned Text or Audio, and one image is inserted at a selected internal segment boundary using the start time of the following segment. Text uses the segment description; Audio synthesizes that description; Video is cropped from the corresponding source-video interval; and Image is extracted at the selected boundary timestamp. The synthesis helper for these segment-level audio assets samples from the four voices used by Audio-to-Motion Generation, with rate offsets of $-20\%$ to $+25\%$ and pitch offsets of $-10$ to $+10$ Hz.

The task JSON stores the ordered components in \texttt{input.interleave.sequence}, together with their temporal intervals or image timestamps. Audio and video are referenced as MP3 and MP4 files, respectively; boundary images are JPEG files. The target is the complete source motion. The 7,764 released tasks contain 33,814 modality components: 8,801 Text, 8,774 Audio, 8,475 Video, and one Image per task.

\subsubsection{Released Task and Asset Organization}\label{sec:release-structure}

The released dataset separates task definitions from referenced assets. \texttt{Tasks/Level1/}, \texttt{Tasks/Level2/}, and \texttt{Tasks/Level3/} contain the task JSON files. Level 1 is organized by task group, task family, and input condition; Level 2 by task family and input condition; Level 3 stores one JSON per ordered multimodal task. File names use the level, full task name, input condition, source-case identifier, and an optional constraint variant. Variants specify repetition counts or alternative ordering instructions.

Input, reference, and target assets are stored under \texttt{Data/Level1/}, \texttt{Data/Level2/}, and \texttt{Data/Level3/}; reusable motions, source videos, rendered skeleton videos, and metadata are stored under \texttt{Data/Shared/}. All asset references are relative to the dataset root. Level 1 and Level 2 task JSON files contain \texttt{metadata}, \texttt{input}, and \texttt{ground\_truth} objects. Within \texttt{input}, \texttt{prompts} stores instructions and \texttt{modalities} stores text and lists of audio, video, image, or spatial inputs. Image+Audio conditions populate both image and audio fields, and the Order Video condition references two input videos in one instance.

Table~\ref{tab:task-conditions} summarizes the released conditions and instance counts. Textual task instructions can accompany non-text conditions; the table lists their substantive conditioning signals. Counts refer to task JSON files.

\begingroup
\small
\begin{longtable}{@{}>{\RaggedRight\arraybackslash}p{0.08\textwidth} >{\RaggedRight\arraybackslash}p{0.30\textwidth} >{\RaggedRight\arraybackslash}p{0.43\textwidth} >{\RaggedLeft\arraybackslash}p{0.10\textwidth}@{}}
\caption{Released task families, input conditions, and benchmark-instance counts.}\label{tab:task-conditions}\\

\toprule
\textbf{Level} & \textbf{Task family} & \textbf{Input conditions} & \textbf{Instances} \\
\midrule
\endfirsthead
\toprule
\textbf{Level} & \textbf{Task family} & \textbf{Input conditions} & \textbf{Instances} \\
\midrule
\endhead
\midrule
\multicolumn{4}{r}{\footnotesize Continued on next page}\\
\endfoot
\bottomrule
\endlastfoot
1 & Text-to-Motion Generation & Text & 38,522 \\
\addlinespace[0.25em]
1 & Video-to-Motion Imitation & Human Video; Skeleton Video & 77,044 \\
\addlinespace[0.25em]
1 & Audio-to-Motion Generation & Audio & 38,522 \\
\addlinespace[0.25em]
1 & Rhythm-to-Motion Alignment & Music + global action summary & 2,552 \\
\addlinespace[0.25em]
1 & Rotation-to-Pose Generation & Joint rotations, root rotation and translation & 38,522 \\
\addlinespace[0.25em]
1 & Motion Prediction & Text; Audio; Human Video; Skeleton Video; Music + summary & 57,619 \\
\addlinespace[0.25em]
1 & Motion Retrodiction & Text; Audio; Human Video; Skeleton Video & 57,332 \\
\addlinespace[0.25em]
1 & Motion Interpolation & Text; Audio; Human Video; Skeleton Video & 31,056 \\
\addlinespace[0.25em]
1 & Key-frame Conditioning & Human/Skeleton Image + Text; Human/Skeleton Image + Audio & 57,332 \\
\addlinespace[0.25em]
1 & Upper-to-Full Body Completion & Text; Audio; Human Video; Skeleton Video & 107,868 \\
\addlinespace[0.25em]
1 & Lower-to-Full Body Completion & Text; Audio; Human Video; Skeleton Video & 107,868 \\
\addlinespace[0.25em]
1 & Target Reaching & Text; Audio (base action + local goal) & 16,386 \\
\addlinespace[0.25em]
2 & Amplitude & Text; Audio; Human Video; Skeleton Video & 10,200 \\
\addlinespace[0.25em]
2 & Speed & Text; Audio; Human Video; Skeleton Video & 9,616 \\
\addlinespace[0.25em]
2 & Direction & Text; Audio; Human Video; Skeleton Video & 20,488 \\
\addlinespace[0.25em]
2 & Body Restrain & Text; Audio; Human Video; Skeleton Video & 13,044 \\
\addlinespace[0.25em]
2 & Order & Text; Audio; two Videos & 47,652 \\
\addlinespace[0.25em]
2 & Times & Text; Audio; Video & 29,214 \\
\addlinespace[0.25em]
2 & Trajectory & Trajectory Image + Text; trajectory Video; Trajectory Image + Audio & 5,115 \\
\addlinespace[0.25em]
3 & Interleaved Multi-Source Steering & Ordered Text, Audio, Video, and boundary Image & 7,764 \\
\addlinespace[0.25em]
\midrule
\multicolumn{3}{r}{\textbf{Total}} & \textbf{773,716} \\
\end{longtable}
\endgroup

\section{Evaluation Details}
\label{Appendix_C}

\subsection{Evaluation Overview}
\label{app:evaluation_overview}

RoboSteer evaluates behavioral steerability by separating basic
behavior-generation capability from the realization of user-specified
steering requirements. Let $\hat{\mathbf M}$ denote the generated
motion, $\mathbf M$ the corresponding reference motion, and $x$ the
input condition. We first evaluate how well $\hat{\mathbf M}$ preserves
the underlying behavior and aligns with the given condition using the
Behavior Generation score, denoted by $\mathrm{BG}$. We then evaluate
whether the generated motion realizes the additional steering
requirements introduced at each level.

For each steering level $k$, we define an Intention Realization score
$\mathrm{IR}_k\in[0,1]$ that quantifies the realization of the
corresponding steering requirement. The Behavior Steering score at
level $l$ is defined uniformly as
\[
\mathrm{BS}_{l}
=
\mathrm{BG}
\prod_{k=1}^{l}
\mathrm{IR}_k,
\qquad
l\in\{1,2,3\}.
\]
Accordingly,
\[
\mathrm{BS}_{\mathrm{Level\,1}}
=
\mathrm{BG}\cdot\mathrm{IR}_1,
\]
\[
\mathrm{BS}_{\mathrm{Level\,2}}
=
\mathrm{BG}\cdot\mathrm{IR}_1\cdot\mathrm{IR}_2,
\]
\[
\mathrm{BS}_{\mathrm{Level\,3}}
=
\mathrm{BG}\cdot\mathrm{IR}_1\cdot\mathrm{IR}_2\cdot\mathrm{IR}_3.
\]

The three realization factors correspond to increasingly complex forms
of behavioral steering. $\mathrm{IR}_1$ measures whether the specified
behavioral condition is realized in Conditional Steering;
$\mathrm{IR}_2$ measures whether the additional execution constraint is
satisfied in Constraint Steering; and $\mathrm{IR}_3$ measures the
realization of temporally interleaved, multi-source steering
requirements in Compositional Steering.

Under this formulation, $\mathrm{BG}$ evaluates basic generation quality,
whereas the $\mathrm{IR}$ factors evaluate whether the steering
requirements at each level are satisfied. The concrete definition and aggregation
of each factor are task-dependent and are detailed in the following
subsections.

\subsection{Behavior Generation Metrics}
\label{app:generation_metrics}

We evaluate basic behavior-generation capability using the conventional
metrics summarized in Section~\ref{sec4.3}. Depending on
the conditioning modality and task, we report FID, Diversity, Contact
Sliding, MM-Distance, Recall@$K$, MPJPE, g-MPJPE,
$E_{\mathrm{vel}}$, BAS-Gen, and BAS-Gap.

Among these metrics, FID, Diversity, MM-Distance, and Recall@$K$ rely
on frozen learned evaluators. FID and Diversity use the OMG Motion
Encoder to represent generated and reference motions in a common motion
feature space. MM-Distance and Recall@$K$ instead use
modality-specific cross-modal evaluators that map the input condition
and generated motion into a shared embedding space. The OMG Motion Encoder is used as a fixed pretrained evaluator, while
the modality-specific cross-modal evaluators are trained for RoboSteer.
Once trained, all evaluators are kept frozen during benchmark evaluation
and use the same parameters for all evaluated models. Their roles are summarized in
Table~\ref{tab:evaluator_summary}.

\begin{table}[h]
    \centering
    \caption{\textbf{Frozen evaluators used for behavior-generation evaluation.}}
    \label{tab:evaluator_summary}

    \small
    \setlength{\tabcolsep}{4pt}
    \renewcommand{\arraystretch}{1.08}

    \begin{tabularx}{0.90\linewidth}{
        @{}
        >{\centering\arraybackslash}p{0.32\linewidth}
        >{\centering\arraybackslash}X
        @{}
    }
        \toprule
        \textbf{Evaluator}
        & \textbf{Evaluation Representation} \\
        \midrule

        OMG Motion Encoder
        & Motion feature space \\

        Text--Motion
        & Shared text--motion space \\

        Audio--Motion
        & Shared audio--motion space \\

        Rhythm--Motion
        & Shared rhythm--motion space \\

        Trajectory--Motion
        & Shared trajectory--motion space \\

        Human Video--Motion
        & Shared human-video--motion space \\

        Skeleton Video--Motion
        & Shared skeleton-video--motion space \\

        \bottomrule
    \end{tabularx}
\end{table}

The remaining metrics do not require learned representation spaces.
Contact Sliding, MPJPE, g-MPJPE, and $E_{\mathrm{vel}}$ are computed
directly from the generated and reference motions, while BAS-Gen and
BAS-Gap measure the temporal correspondence between motion and rhythm
signals.

\paragraph{Motion Representation and Temporal Processing.}
All generated and reference motions are converted to a unified G1
representation. Each frame is represented as
\[
\mathbf q_t
=
[
\mathbf r_t,\,
\mathbf u_t,\,
\boldsymbol{\theta}_t
]
\in\mathbb R^{36},
\]
where $\mathbf r_t\in\mathbb R^3$ denotes the root translation,
$\mathbf u_t\in\mathbb R^4$ denotes the root orientation represented
as a quaternion in $(w,x,y,z)$ order, and
$\boldsymbol{\theta}_t\in\mathbb R^{29}$ contains the articulated
joint angles. Root quaternions are normalized and made sign-continuous
along the temporal dimension. When different motion fields contain
unequal numbers of frames, their minimum common frame count is used.

Different metrics use temporal processing appropriate to their
evaluation objective. For FID and Diversity, each complete motion is
independently resampled to 60 endpoint-inclusive phase samples. Root
translations and joint angles are linearly interpolated, while root
rotations use sign-consistent normalized linear interpolation.
Metrics that depend on physical execution time retain the original
$50$ Hz temporal scale. For a motion containing $N$ valid frames, its
duration is
\[
d=\frac{N-1}{50}.
\]
Contact Sliding is evaluated directly at this temporal resolution. For
MPJPE, g-MPJPE, and $E_{\mathrm{vel}}$, the generated motion is
temporally aligned to the effective reference sequence before
frame-wise comparison.

\paragraph{Behavior Generation Score.}
The Behavior Generation score combines two complementary aspects of
generation quality: similarity to the reference behavior and alignment
with the input condition. We use FID for generated--reference
consistency and the modality-specific MM-Distance for
generated--condition consistency:
\[
\mathrm{BG}
=
1000
\exp\!\left(
-0.3\,\mathrm{FID}
-1.6\,\mathrm{MM\text{-}Distance}
\right).
\]

Both FID and MM-Distance are lower-is-better metrics. The exponential
transformation therefore maps smaller discrepancies to higher BG
scores. The coefficients $0.3$ and $1.6$ are empirically calibrated to
balance the contributions of the two metrics, while the factor $1000$
only rescales the score for convenient reporting.

For each conditioning modality, BG combines the common FID with the
MM-Distance produced by the corresponding cross-modal evaluator in
Table~\ref{tab:evaluator_summary}. It therefore provides the common
behavior-generation component used by the Behavior Steering scores at
all three steering levels.

\subsection{Level 1: Conditional Steering}
\label{app:level1_metrics}

Level~1 evaluates whether the generated motion faithfully realizes the
provided behavioral condition under different degrees of spatiotemporal
completeness. Its Behavior Steering score is defined as
\[
\mathrm{BS}_{\mathrm{Level\,1}}
=
\mathrm{BG}\cdot\mathrm{IR}_1,
\]
where $\mathrm{IR}_1\in[0,1]$ denotes the Intention Realization score
for Conditional Steering. Since different Level~1 tasks specify the
behavioral condition in different forms, $\mathrm{IR}_1$ is computed
using task-specific evaluation rules.

\paragraph{Full Conditioning Reproduction.}
For full-conditioning tasks, the behavioral condition is fully specified
by the input, so no additional completion requirement is introduced. We
therefore set
\[
\mathrm{IR}_1 = 1,
\]
and the Level~1 Behavior Steering score reduces to
\[
\mathrm{BS}_{\mathrm{Level\,1}}=\mathrm{BG}.
\]

\paragraph{Temporal Completion.}
Motion Prediction, Motion Retrodiction, and Motion Interpolation provide
only partial temporal observations and require the model to complete the
remaining motion. We evaluate whether the generated motion recovers the
expected temporal extent of the complete behavior using Temporal
Intersection-over-Union (\textbf{tIoU}).

For sample $n$, let $\hat d_n$ and $d_n$ denote the durations of the
generated and reference motions, respectively, with temporal intervals
\[
\hat{\mathcal I}_n=[0,\hat d_n],
\qquad
\mathcal I_n=[0,d_n].
\]
We define
\[
\mathrm{tIoU}_n
=
\frac{
\left|
\hat{\mathcal I}_n\cap\mathcal I_n
\right|
}{
\left|
\hat{\mathcal I}_n\cup\mathcal I_n
\right|
}
=
\frac{
\min(\hat d_n,d_n)
}{
\max(\hat d_n,d_n)
}.
\]
The Intention Realization score is then
\[
\mathrm{IR}_1
=
\frac{1}{|\mathcal D|}
\sum_{n\in\mathcal D}
\mathrm{tIoU}_n,
\]
where $\mathcal D$ denotes the set of valid samples. A score of one
indicates identical temporal extent, while the score decreases as the
generated duration deviates from the reference duration. The same rule
is used for Motion Prediction, Motion Retrodiction, and Motion
Interpolation.

\paragraph{Spatial Completion.}
Upper-to-Full and Lower-to-Full Body Completion provide either the
upper-body or lower-body motion as a partial condition and require the
model to complete the full-body motion. We evaluate only the body region
that is missing from the input using Joint-Range Overlap
(\textbf{JRO}), which measures the overlap between generated and
reference joint-angle ranges.

For joint $j$ in sample $n$, let
\[
\hat R_{n,j}
=
[\hat a_{n,j},\hat b_{n,j}]
=
\left[
\min_t\hat\theta_{n,t,j},
\max_t\hat\theta_{n,t,j}
\right]
\]
and
\[
R_{n,j}
=
[a_{n,j},b_{n,j}]
=
\left[
\min_t\theta_{n,t,j},
\max_t\theta_{n,t,j}
\right]
\]
denote the generated and reference joint-angle ranges. Their overlap
length is
\[
o_{n,j}
=
\max\left(
0,\,
\min(\hat b_{n,j},b_{n,j})
-
\max(\hat a_{n,j},a_{n,j})
\right).
\]
We define the normalization term as
\[
z_{n,j}
=
\begin{cases}
\hat b_{n,j}-\hat a_{n,j},
& \hat R_{n,j}\supseteq R_{n,j},\\
b_{n,j}-a_{n,j},
& \text{otherwise},
\end{cases}
\]
and compute
\[
\mathrm{JRO}_{n,j}
=
\frac{o_{n,j}}{z_{n,j}}.
\]
This formulation penalizes both insufficient motion range and excessive
expansion beyond the reference range.

When both generated and reference ranges are nearly static, the overlap
ratio becomes ill-defined. If both range widths are at most
$10^{-4}$ rad, we instead compare their range centers:
\[
\mathrm{JRO}_{n,j}
=
\mathbf{1}\left[
\left|
\frac{\hat a_{n,j}+\hat b_{n,j}}{2}
-
\frac{a_{n,j}+b_{n,j}}{2}
\right|
\leq 0.05\ \mathrm{rad}
\right].
\]

For Upper-to-Full, the evaluated missing region consists of lower-body
joint degrees of freedom $0$--$11$. For Lower-to-Full, the evaluated
region consists of waist and upper-body degrees of freedom $12$--$28$.
For each sample, JRO is first averaged over the corresponding missing
joint set, and $\mathrm{IR}_1$ is then macro-averaged over valid samples:
\[
\mathrm{IR}_1
=
\frac{1}{|\mathcal D|}
\sum_{n\in\mathcal D}
\frac{1}{|\mathcal J_n|}
\sum_{j\in\mathcal J_n}
\mathrm{JRO}_{n,j},
\]
where $\mathcal J_n$ denotes the evaluated missing-body joint set.

\paragraph{Key-Frame Conditioning.}
Key-Frame Conditioning evaluates whether the generated motion satisfies
a set of sparsely specified poses at annotated timestamps. For each
key-frame timestamp $\tau_k$, we interpolate the reference pose at
$\tau_k$ and evaluate the generated motion within a short temporal
window centered at that timestamp:
\[
\mathcal W_k
=
\{
\tau_k-0.04,\,
\tau_k-0.02,\,
\tau_k,\,
\tau_k+0.02,\,
\tau_k+0.04
\}.
\]

Pose comparison is performed after G1 forward kinematics using the fixed
14-link subset
\[
\{0,4,10,18,5,11,19,9,16,22,28,17,23,29\}.
\]
For each candidate time $r\in\mathcal W_k$, we remove pelvis translation
independently from the generated and reference poses and compute the mean
Euclidean distance between corresponding 3D link positions. Let
$e_{k,r}$ denote this pelvis-aligned pose error. We select the best
temporal match as
\[
e_k
=
\min_{r\in\mathcal W_k} e_{k,r},
\]
and convert it into a normalized matching score:
\[
s_k
=
\exp\left(
-\frac{e_k}{0.20\ \mathrm{m}}
\right).
\]
The Level~1 Intention Realization score is computed by averaging over
all valid annotated key frames:
\[
\mathrm{IR}_1
=
\frac{1}{K}
\sum_{k=1}^{K}
s_k.
\]
Candidate timestamps outside the valid task interval are clamped to the
corresponding temporal boundary. The temporal window allows small timing
deviations while still requiring the generated pose to closely match
the specified key frame.

\paragraph{Target Reaching.}
Target Reaching evaluates whether the generated motion reaches a set of
required target poses throughout the behavior. We use four reference
pose anchors located at $25\%$, $50\%$, $75\%$, and $100\%$ of the task
duration. Each target is evaluated using the same forward-kinematics
representation, pelvis alignment, temporal interpolation, and
five-frame matching window as in Key-Frame Conditioning.

For sample $n$ and target $k$, let
\[
e_{n,k}
=
\min_{r\in\mathcal W_{n,k}}
e_{n,k,r}
\]
denote the minimum pelvis-aligned pose error within the corresponding
temporal window. The Intention Realization score is
\[
\mathrm{IR}_1
=
\frac{1}{|\mathcal D|}
\sum_{n\in\mathcal D}
\frac{1}{4}
\sum_{k=1}^{4}
\exp\left(
-\frac{e_{n,k}}{0.20\ \mathrm{m}}
\right).
\]

Together, these task-specific rules instantiate $\mathrm{IR}_1$
according to the form of the input condition: temporal overlap for
Temporal Completion, joint-range overlap for Spatial Completion, and
pose matching for sparse key-frame and target conditions.

\subsection{Level 2: Constraint Steering}
\label{app:level2_metrics}

Level~2 evaluates whether a model can satisfy an additional behavioral
constraint while preserving the underlying behavior specified at
Level~1. Its Behavior Steering score is defined as
\[
\mathrm{BS}_{\mathrm{Level\,2}}
=
\mathrm{BG}\cdot\mathrm{IR}_1\cdot\mathrm{IR}_2,
\]
where $\mathrm{IR}_2\in[0,1]$ denotes the Intention Realization score
for Constraint Steering.

RoboSteer considers seven constraint families: Speed, Amplitude,
Direction, Order, Times, Trajectory, and Body Restrain. For constraint
family $c$, let $b_{n,c}\in\{0,1\}$ indicate whether the generated
motion of sample $n$ satisfies the corresponding constraint. We define
the family-level Constraint Grounding score as
\[
\mathrm{CG}_c
=
\frac{1}{|\mathcal D_c|}
\sum_{n\in\mathcal D_c}
b_{n,c},
\]
where $\mathcal D_c$ denotes the set of valid samples for family $c$.
The Level~2 Intention Realization score is obtained by macro-averaging
over the seven families:
\[
\mathrm{IR}_2
=
\frac{1}{7}
\sum_{c=1}^{7}
\mathrm{CG}_c.
\]
\paragraph{Speed.}
Speed evaluates whether the generated motion changes its execution
duration in the requested direction. Let $\hat d_n$ and $d_n$ denote
the durations of the generated and reference motions, respectively.
Constraint satisfaction is defined as
\[
b_{n,\mathrm{speed}}
=
\begin{cases}
\mathbf{1}[\hat d_n < d_n],
& \texttt{fast},\\
\mathbf{1}[\hat d_n > d_n],
& \texttt{slow}.
\end{cases}
\]

\paragraph{Amplitude.}
Amplitude evaluates whether the overall motion magnitude is increased
or decreased as requested. For a motion $\mathbf M$, we define its
global amplitude as the maximum joint-angle range over all 29 articulated
joints:
\[
A(\mathbf M)
=
\max_{1\leq j\leq 29}
\left(
\max_t\theta_{t,j}
-
\min_t\theta_{t,j}
\right).
\]
The generated motion satisfies the amplitude constraint when
\[
b_{n,\mathrm{amp}}
=
\begin{cases}
\mathbf{1}
\left[
A(\hat{\mathbf M}_n)
>
A(\mathbf M_n)
\right],
& \texttt{scale\_up},\\
\mathbf{1}
\left[
A(\hat{\mathbf M}_n)
<
A(\mathbf M_n)
\right],
& \texttt{scale\_down}.
\end{cases}
\]

\paragraph{Direction.}
Direction evaluates whether the generated motion moves toward the
requested lateral direction relative to the robot's initial heading.
Let
\[
\Delta\mathbf r^{xy}
=
\hat{\mathbf r}^{xy}_{T}
-
\hat{\mathbf r}^{xy}_{1}
=
(\Delta r_x,\Delta r_y)
\]
denote the horizontal root displacement between the first and last
frames. We define
\[
\mathbf f_0=(f_{0,x},f_{0,y})
\]
as the horizontal projection of the root's local $+X$ axis at the
first frame and use it as the initial forward direction.

The heading-relative displacement angle is
\[
\phi
=
\operatorname{atan2}
\left(
f_{0,x}\Delta r_y-f_{0,y}\Delta r_x,\,
f_{0,x}\Delta r_x+f_{0,y}\Delta r_y
\right).
\]
Under this convention, the target angles are
\[
\phi^\star=
\begin{cases}
+90^\circ, & \texttt{left},\\
-90^\circ, & \texttt{right}.
\end{cases}
\]
We allow a tolerance of $60^\circ$ and additionally require a minimum
horizontal displacement of $0.10$ m. The binary score is therefore
\[
b_{n,\mathrm{dir}}
=
\mathbf{1}
\left[
\left\|
\Delta\mathbf r^{xy}
\right\|_2
\geq
0.10\ \mathrm{m}
\right]
\cdot
\mathbf{1}
\left[
\left|
\operatorname{wrap}
(\phi-\phi^\star)
\right|
\leq
60^\circ
\right],
\]
where $\operatorname{wrap}(\cdot)$ maps the angular difference to the
principal angular interval. The displacement threshold prevents
nearly stationary motions from satisfying the constraint through
numerical noise.

\paragraph{Order.}
Order evaluates whether two requested actions are executed in the
specified temporal order. We render each generated motion using a
standardized rendering configuration and evaluate the resulting video
with a fixed VideoLLM.

For each sample, the evaluator receives the complete rendered motion
video together with two sample-specific candidate actions, denoted by
$A_n$ and $B_n$. The required temporal order is not provided to the
evaluator. 

Let $\hat v_n^A$ and $\hat v_n^B$ indicate whether the evaluator
recognizes the two candidate actions as visibly and completely
performed. Let
\[
\widehat F_n
\in
\{A,B,\mathrm{simultaneous},\mathrm{unclear}\}
\]
denote which action is recognized as beginning first, and let
$F_n^\star\in\{A,B\}$ denote the target first action. We define
\[
b_{n,\mathrm{order}}
=
\mathbf{1}
\left[
\hat v_n^A=1
\land
\hat v_n^B=1
\land
\widehat F_n=F_n^\star
\right].
\]
Thus, a sample satisfies the Order constraint only when both requested
actions are visibly executed and their onset order is correct. A sample
is treated as incorrect if either action is absent or incomplete, if the
two actions begin simultaneously, or if their temporal order is unclear.
Incidental actions and transitions are ignored.

\paragraph{Times.}
Times evaluates whether a target action is visibly performed the
requested number of times. It is evaluated independently from Order
using a task-specific VideoLLM prompt.

For each sample, the evaluator receives the complete rendered motion
video together with the sample-specific target action $A_n$. The
required repetition count $N_n^\star$ is not provided to the evaluator.
The evaluator predicts a visibility indicator $\hat v_n$ and the number
of complete visible instances $\hat N_n(A_n)$. A continuously sustained
action is counted once, whereas an action that terminates or resets and
is subsequently performed again is counted as a new instance. Incomplete
instances at the beginning or end of the video are not counted.

The predicted count is compared with the hidden target count using the
fixed Times count-matching rule:
\[
b_{n,\mathrm{times}}
=
\mathbf{1}
\left[
\hat v_n=1
\land
M_{\mathrm{times}}
\bigl(\hat N_n(A_n),N_n^\star\bigr)=1
\right],
\]

\paragraph{VideoLLM Evaluation Protocol.}
Order and Times are evaluated using the same fixed VideoLLM checkpoint,
standardized motion-video rendering and sampling procedure, and
deterministic decoding configuration. They use separate task-specific
prompts and output schemas.

Action identities are taken from the frozen task manifests. For Order,
the evaluator is given the two candidate actions but not their target
temporal order. For Times, it is given the target action but not the
required repetition count. For modalities without textual action labels,
such as audio and video, the action identities are inherited from the
corresponding paired text-task records.

The fixed Order prompt is:

\begin{promptbox}[VideoLLM Order Prompt]

You are evaluating a rendered human/humanoid motion video.

Candidate action A: \texttt{<ACTION\_A>}

Candidate action B: \texttt{<ACTION\_B>}

Based only on the visible motion, determine whether each candidate
action is performed. If both are performed, determine which action
begins first.

\textbf{Rules:}
\begin{enumerate}
    \item Do not infer an intended order from the candidate names or display order.
    \item Mark an action visible only when its characteristic motion is clearly executed.
    \item Determine order from action onset, not peak-motion or completion time.
    \item Ignore incidental standing, balancing, locomotion, and transitions.
    \item Return \texttt{simultaneous} if the two actions begin together.
    \item Return \texttt{unclear} if their order cannot be determined reliably.
    \item Return only valid JSON.
\end{enumerate}

\texttt{\{"a\_visible": true, "b\_visible": true,}\\
\texttt{"first": "A" | "B" | "simultaneous" | "unclear"\}}

\end{promptbox}

The fixed Times prompt is:

\begin{promptbox}[VideoLLM Times Prompt]

You are evaluating a rendered human/humanoid motion video.

Target action: \texttt{<ACTION>}

Based only on the visible motion, count the number of complete instances
of the target action.

\textbf{Rules:}
\begin{enumerate}
    \item The requested count is not provided.
    \item Count only clearly and completely executed target actions.
    \item A continuously sustained action counts once.
    \item A new instance requires the previous instance to end or reset.
    \item For cyclic actions, count complete cycles rather than frames or intermediate poses.
    \item An incomplete instance at the beginning or end does not count.
    \item Repetitions of unrelated motions do not count.
    \item Return only valid JSON.
\end{enumerate}

\texttt{\{"visible": true, "count": 0 | 1 | 2 | 3 | 4 | 5 | "5+"\}}

\end{promptbox}

The same VideoLLM checkpoint, rendering configuration, action vocabulary,
prompt template, and decoding settings are used for all evaluated
models. Outputs that cannot be parsed into the prescribed JSON format
or contain labels outside $\mathcal A_{\mathrm{RS}}$ are treated as
invalid recognition outputs.

\paragraph{Trajectory.}
Trajectory evaluates whether the generated root trajectory follows the
requested spatial pattern. Let
\[
\Gamma_n
=
\left\{
\hat{\mathbf r}^{xy}_{n,t}
\right\}_{t=1}^{T}
\]
denote the generated root trajectory projected onto the ground plane.

Before computing trajectory geometry, we smooth the projected path using
a five-frame edge-padded moving average while preserving its two
endpoints. Consecutive points separated by less than
\[
\max(0.005\ \mathrm{m},\,0.01D_{\max})
\]
are removed to reduce sensitivity to small numerical oscillations.
From the resulting path, we compute the maximum displacement
$D_{\max}$, path length $L$, start--end distance $D_{\mathrm{end}}$,
normalized enclosed area $A_N$, and cumulative heading change $\Theta$.

All trajectory constraints require sufficient overall displacement:
\[
D_{\max}\geq0.25\ \mathrm{m}.
\]

For a back-and-forth trajectory, the generated path must satisfy
\[
\frac{D_{\mathrm{end}}}{D_{\max}}\leq0.25,
\qquad
\frac{L}{D_{\max}}\geq1.60,
\]
\[
\frac{P_+}{D_{\max}}\geq0.30,
\qquad
\frac{P_-}{D_{\max}}\geq0.30,
\qquad
A_N\leq0.05,
\]
where $P_+$ and $P_-$ denote accumulated travel along the positive and
negative directions of the principal trajectory axis.

For a circular trajectory, the path must satisfy
\[
A_N\geq0.10,
\qquad
\Theta\geq\pi.
\]

For an S-shaped trajectory, four approximately arc-length-equidistant
landmarks are extracted from the generated path. The three consecutive
segments must exhibit alternating lateral turns
(left--right--left or right--left--right), with each turn exceeding
$15^\circ$. The lateral landmark range must additionally be at least
$0.15D_{\max}$, and
\[
\frac{D_{\mathrm{end}}}{D_{\max}}
\geq0.35.
\]
The binary score $b_{n,\mathrm{traj}}$ is set to one only when all
requirements associated with the requested trajectory pattern are
satisfied.

\paragraph{Body Restrain.}
Body Restrain evaluates whether a specified body region remains
sufficiently stationary while the rest of the body continues to move.
This prevents the constraint from being satisfied trivially by making
the entire body static.

We first apply G1 forward kinematics to obtain 3D body-link positions.
The positions are transformed into root-aligned coordinates to remove
motion caused by global root translation and rotation. Link velocities
are then computed at $50$ Hz.

For a set of links $\mathcal J$, we define its motion activity as the
root-mean-square link velocity
\[
V(\mathcal J)
=
\sqrt{
\frac{1}{(T-1)|\mathcal J|}
\sum_{t=1}^{T-1}
\sum_{j\in\mathcal J}
\left\|
\mathbf v_{t,j}
\right\|_2^2
}.
\]

Let $\mathcal J_{\mathrm{restricted}}$ denote the restrained links and
$\mathcal J_{\mathrm{remaining}}$ the remaining links. When restraining
the arms, links $16$--$29$ form the restricted set and links $1$--$15$
form the remaining set. When restraining the legs, links $1$--$12$ are
restricted and links $13$--$29$ form the remaining set. The root is
excluded from the activity computation.

A generated motion satisfies the constraint when
\[
b_{n,\mathrm{res}}
=
\mathbf{1}
\left[
V(\mathcal J_{\mathrm{restricted}})
\leq
0.10\ \mathrm{m/s}
\right]
\cdot
\mathbf{1}
\left[
V(\mathcal J_{\mathrm{remaining}})
\geq
0.12\ \mathrm{m/s}
\right].
\]
The first condition requires the specified body region to remain
sufficiently still, while the second ensures that the remaining body
continues to exhibit non-trivial motion.

Together, the seven family-level Constraint Grounding scores measure
how reliably a model follows explicit execution constraints while
preserving the underlying behavior.

\subsection{Level 3: Compositional Steering}
\label{app:level3_metrics}

Level~3 evaluates whether a model can consistently follow multiple
modality-specific steering requirements that are temporally interleaved
within a single motion. Each Level~3 sample is divided into multiple
temporal segments, with each segment associated with one steering
requirement.

For the $i$-th temporal segment, let $\Delta t_i$ denote its duration
and let $s_i\in[0,1]$ denote the realization score of the steering
requirement associated with that segment. We aggregate the segment-level
scores according to their durations:
\[
\mathrm{IR}_3
=
\frac{
\sum_{i=1}^{K}
\Delta t_i s_i
}{
\sum_{i=1}^{K}
\Delta t_i
},
\]
where $K$ denotes the number of temporal segments.

Following the unified formulation in
Section~\ref{app:evaluation_overview}, the Level~3 Behavior Steering
score is
\[
\mathrm{BS}_{\mathrm{Level\,3}}
=
\mathrm{BG}
\cdot
\mathrm{IR}_1
\cdot
\mathrm{IR}_2
\cdot
\mathrm{IR}_3.
\]

The duration-weighted aggregation gives each segment a contribution
proportional to the time span over which its steering requirement is
active. In this way, Level~3 focuses specifically on whether a model can
maintain coherent behavior while switching among different
modality-specific requirements over time, without conflating this
capability with the lower-level completion and explicit-constraint
failures evaluated separately in Level~1 and Level~2.

\section{Experiment Detail}
\label{Appendix_D}

\subsection{Data Preprocessing}

We develop a hierarchical preprocessing pipeline that converts heterogeneous text, audio, and image conditions from Levels~1--3 into a unified representation compatible with downstream motion generation models, enabling consistent evaluation across modalities and task settings.

\paragraph{Level-1 data.}
Since downstream motion generation models do not directly accept audio or image inputs, we first convert these modalities into supported representations. We transcribe speech using \textbf{faster-whisper}. For image-conditioned tasks with timestamped key frames, we construct static videos by holding each frame until the timestamp of the subsequent frame. This preserves the original visual evidence and temporal structure without introducing artificial motion.

For generation and spatial completion tasks, existing motion generation models cannot reliably process long and information-rich textual instructions. We therefore use \textbf{Qwen2.5-3B-Instruct} to rewrite the original descriptions as concise English imperative motion instructions. This normalization removes scene descriptions, camera-related details, conversational expressions, repetitions, and other non-motion content, while preserving actions, interacted objects, directions, temporal order, key body-state constraints, and necessary object interactions.

For predictive text tasks, including future prediction, motion retrospection, and motion interpolation, we apply the same textual normalization to the observed motion descriptions and represent the unobserved interval using a single task-specific prediction prompt. Specifically, we append \textit{Generate smooth future motion} to the observed prefix for future prediction, prepend \textit{Generate smooth preceding motion} to the observed suffix for retrospection, and insert \textit{Generate smooth transition motion} between the preceding and following observed segments for interpolation.

\paragraph{Level-2 data.}
We separately process modifier-based and trajectory-based descriptions. Modifier-based tasks include motion amplitude, direction, body restraint, and speed. Since downstream models cannot jointly process an original motion description and an additional constraint as separate inputs, we use \textbf{Qwen2.5-3B-Instruct} to combine them into a standalone English imperative instruction. We further enforce category-specific semantic constraints to ensure that the target attribute is modified while irrelevant attributes remain unchanged. For example, direction editing removes descriptions that conflict with the target direction; body-restraint editing explicitly specifies that constrained limbs remain still and removes infeasible actions involving these limbs; and amplitude and speed editing modify only the designated attribute while preserving the original action semantics and direction. For trajectory-based tasks, we compress verbose descriptions into instructions containing at most two high-level action clauses, while preserving temporal order, directional cues, and necessary object interactions.

\paragraph{Level-3 data.}
We unify text annotations and audio transcriptions through a shared manifest, and use \textbf{Qwen2.5-3B-Instruct} to normalize them into single-line English imperative instructions. The process removes non-motion content while retaining actions, interacted objects, directional information, and temporal order, thereby providing a consistent textual representation for text- and audio-derived conditions.

To reduce noise introduced by automatic processing, we apply rule-based validation and iterative refinement across all levels. We verify output format, text length, temporal consistency, and task-specific semantic constraints. Samples that violate these constraints are regenerated or deterministically corrected when possible; unrecoverable samples are excluded from the final dataset. This pipeline converts heterogeneous task data into unified conditions accepted by downstream models, allowing consistent and comparable evaluation under diverse modalities and task settings.

\subsection{Model Adaptation and Inference Protocol}

The preprocessing stage provides a unified condition representation, whereas each baseline model retains its native input interface, inference procedure, and output parameterization. We therefore adapt each model at the interface level without modifying its pretrained weights. Model-specific configuration files, checkpoints, inference commands, input manifests, random seeds, output directories, and validation results are documented separately for reproducibility.

\paragraph{Text-conditioned models.}
For \textbf{GEM}, we feed the normalized English imperative instructions directly into its text-conditioned generation interface. \textbf{Language of Motion (LoM)} treats each non-empty text line as an independent prompt and generates a corresponding motion segment. \textbf{MotionCraft} concatenates the retained descriptions within each text file into a single prompt before generation. \textbf{TextOp} employs its native \textbf{RobotMDAR} high-level autoregressive module to generate text-conditioned robot-motion references, which are subsequently tracked by its low-level motion-tracking policy. The original 23-DoF RobotMDAR output is subsequently expanded to the unified 29-DoF representation. \textbf{UniAct} uses its native text-conditioned control interface and fixed-rate motion generation procedure. \textbf{UH-1} uses its original text-conditioned inference pipeline, followed by a dedicated cross-embodiment conversion from H1 to G1.

\paragraph{Predictive text tasks.}
For predictive tasks, \textbf{GEM}, \textbf{TextOp}, \textbf{UniAct}, and \textbf{UH-1} process the complete normalized input, including observed descriptions and the prediction prompt, and retain only the motion segment corresponding to the prediction instruction. This preserves available context while ensuring that evaluation is performed on the unobserved interval. In contrast, \textbf{LoM} generates each text line independently; to reduce computation, we directly generate the segment associated with the prediction instruction. \textbf{MotionCraft} merges multiple text lines into a single prompt and does not expose instruction-level motion boundaries; therefore, it cannot provide a directly comparable prediction-only segment.

\paragraph{Audio-conditioned models.}
For \textbf{Listen, Denoise, Action! (LDA)}, we use the model's native audio-conditioned inference interface. For \textbf{MotionCraft}, we use its music-to-dance branch, which extracts 30~Hz, 35-dimensional music features from each input audio sequence, including onset, MFCC, chroma, beat, and rhythm-related features. A fixed generic text prompt is retained only to satisfy the model interface; no sample-specific textual condition is provided. Therefore, this setting evaluates music-conditioned motion generation rather than joint text--music conditioning.

\paragraph{Video- and image-conditioned models.}
For \textbf{GEM}, preprocessed static videos are passed through its native video-conditioned interface. For \textbf{video2robot}, the visual condition is processed through the original image/video encoder and subsequently converted into robot motion using the released inference pipeline.

\paragraph{Reference-motion model.}
For \textbf{BFM-Zero}, we use the provided reference motion as input to its simulator-based generation procedure. The resulting states are exported directly from simulation and converted into the common evaluation representation.

For all methods, failed samples are explicitly recorded rather than silently replaced. We distinguish model failures, invalid outputs, conversion failures, and incomplete generations from successfully exported samples, ensuring that final evaluation statistics reflect the actual runnable subset of each baseline.

\subsection{Post-processing and Unified Motion Representation}

To enable consistent evaluation across models with different embodiments and output formats, we convert all valid outputs into a unified SONIC representation based on the 29-DoF G1 joint order used by IsaacLab. Each exported sequence contains root states, joint states, and the positions, orientations, linear velocities, and angular velocities of 14 key rigid bodies. The reduced set of rigid bodies is sufficient to capture the major kinematic structure required for motion evaluation while avoiding redundant simulator-specific links.

\paragraph{Human-motion generation models.}
Outputs from human-motion generation models, including \textbf{GEM}, \textbf{LoM}, \textbf{MotionCraft}, \textbf{video2robot}, and \textbf{LDA}, are processed using the General Motion Retargeting (GMR) pipeline. Specifically, GMR retargets human motion to the G1 embodiment, resamples the resulting trajectory to 50~Hz, converts the motion from MuJoCo to IsaacLab coordinates, and recovers rigid-body states through forward kinematics. It further applies a temporally smooth ground-alignment procedure based on foot collision geometry, correcting only substantial vertical penetration while preserving airborne motion whenever possible. Joint velocities, rigid-body linear velocities, and angular velocities are recomputed from the final exported states to ensure consistency with the final coordinate system, time step, and kinematic configuration.

\paragraph{Simulator-native outputs.}
\textbf{BFM-Zero} directly exports simulator states that are already consistent with the target representation. We therefore preserve its state sequence and only convert quaternion ordering from \texttt{xyzw} to \texttt{wxyz} when required. Similarly, the \textbf{TextOp} pipeline produces robot motion that can be directly mapped to the SONIC representation. Its motion sequences are retained without temporal resampling, while missing G1 joints are filled according to the established joint mapping.

\paragraph{UniAct outputs.}
\textbf{UniAct} exports motion at 50~Hz and therefore does not require temporal resampling. We preserve its original frame sequence, root trajectory, and joint velocities, and recover rigid-body states through forward kinematics. Rigid-body linear and angular velocities are then computed from the reconstructed body states to ensure consistency with the exported representation. A smooth vertical translation based on the lowest reconstructed body height is applied to reduce noticeable ground penetration without altering horizontal motion or root orientation.

\paragraph{UH-1 outputs.}
For \textbf{UH-1}, we map the 27-DoF H1 joint states into the 29-DoF G1 joint space. The two G1 waist degrees of freedom without corresponding H1 joints are set to zero, and the remaining joints are clipped to the valid G1 physical limits. We preserve the original horizontal root trajectory and heading, resample root and joint states to 50~Hz, and recover rigid-body states through G1 forward kinematics. A smooth vertical ground-alignment correction is then applied to reduce penetration while preserving horizontal movement and root orientation. Velocities are recomputed from the final exported states.

Finally, we validate all exported sequences for joint dimensions, frame consistency, quaternion normalization, finite values, and compatibility with the SONIC schema. Invalid sequences are excluded from evaluation. This unified post-processing procedure enables fair comparison among human-motion, robot-motion, audio-conditioned, text-conditioned, image-conditioned, and simulator-based methods under a shared robot embodiment and state representation.

\section{Detailed Experimental Supplement}
\label{Appendix_E}

\begin{sidewaystable}[p]
\centering
\begingroup
\scriptsize
\setlength{\tabcolsep}{2.3pt}
\renewcommand{\arraystretch}{1.12}
\arrayrulecolor{black}

\resizebox{\textwidth}{!}{%
\begin{tabular}{l|*{5}{c}|*{4}{c}|*{3}{c}}
\toprule

\multicolumn{1}{c|}{}
& \multicolumn{5}{c|}{\textbf{Full Conditioning Reproduction}}
& \multicolumn{4}{c|}{\textbf{Temporal Completion}}
& \multicolumn{3}{c}{\textbf{Spatial Completion}}
\\

\cmidrule(lr){2-6}
\cmidrule(lr){7-10}
\cmidrule(lr){11-13}

\textbf{Model}
& \textbf{T2M}
& \textbf{A2M}
& \textbf{V2M}
& \textbf{R2M}
& \textbf{R2P}
& \textbf{Pred.}
& \textbf{Retro.}
& \textbf{Inter.}
& \textbf{KC}
& \textbf{U2F}
& \textbf{L2F}
& \textbf{TR}
\\

\midrule

% ============================================================
% Text
% ============================================================
\rowcolor{textblue}
\multicolumn{13}{l}{\textbf{Text}}
\\

GEM + GMR
& \bgbsfmt{\second{115.659}}{\second{115.659}}
& -- & -- & -- & --
& \bgbsfmt{\second{107.224}}{\second{74.059}}
& \bgbsfmt{\best{105.749}}{\best{73.041}}
& \bgbsfmt{\best{109.635}}{\best{72.075}}
& \bgbs{113.329}{33.307}
& \bgbsfmt{\second{115.271}}{\second{55.856}}
& \bgbsfmt{\second{118.343}}{\second{49.474}}
& \bgbs{113.392}{33.529}
\\

LoM + GMR
& \bgbs{103.900}{103.900}
& -- & -- & -- & --
& \bgbs{98.019}{61.309}
& \bgbs{88.546}{49.951}
& \bgbs{90.525}{53.124}
& \bgbsfmt{\best{117.348}}{\best{48.495}}
& \bgbsfmt{\best{106.972}}{\best{56.860}}
& \bgbsfmt{\best{106.703}}{\best{58.981}}
& \bgbsfmt{\best{118.176}}{\best{48.552}}
\\

MotionCraft + GMR
& \bgbsfmt{\best{116.900}}{\best{116.900}}
& -- & -- & -- & --
& \bgbsfmt{\best{109.138}}{\best{75.348}}
& \bgbsfmt{\second{105.606}}{\second{72.910}}
& \bgbsfmt{\second{106.363}}{\second{70.545}}
& \bgbsfmt{\second{117.959}}{\second{47.536}}
& \bgbs{111.343}{33.263}
& \bgbs{123.782}{43.012}
& \bgbsfmt{\second{110.140}}{\second{45.236}}
\\

TextOp
& \bgbs{80.194}{80.194}
& -- & -- & -- & --
& \bgbs{78.145}{41.059}
& \bgbs{75.528}{39.684}
& \bgbs{80.407}{40.200}
& \bgbs{98.292}{35.625}
& \bgbs{77.719}{32.986}
& \bgbs{88.294}{26.265}
& \bgbs{86.440}{30.118}
\\

UniAct
& \bgbs{104.924}{104.924}
& -- & -- & -- & --
& \bgbs{102.250}{40.112}
& \bgbs{85.494}{54.621}
& \bgbs{89.061}{44.539}
& \bgbs{106.932}{36.606}
& \bgbs{100.141}{52.910}
& \bgbs{108.918}{39.101}
& \bgbs{105.989}{36.524}
\\

UH-1
& \bgbs{97.650}{97.650}
& -- & -- & -- & --
& \bgbs{78.047}{47.609}
& \bgbs{67.214}{43.355}
& \bgbs{77.925}{50.507}
& \bgbs{115.878}{38.477}
& \bgbs{96.276}{38.845}
& \bgbs{108.570}{19.526}
& \bgbs{105.733}{35.239}
\\

\midrule

% ============================================================
% Audio
% ============================================================
\rowcolor{audiogreen}
\multicolumn{13}{l}{\textbf{Audio}}
\\

GEM + GMR
& --
& \bgbs{97.413}{97.413}
& -- & -- & --
& \bgbsfmt{\second{97.237}}{\second{67.161}}
& \bgbsfmt{\second{96.518}}{\second{66.664}}
& \bgbsfmt{\second{100.632}}{\second{66.156}}
& \bgbs{105.813}{31.098}
& \bgbs{100.434}{48.666}
& \bgbsfmt{\second{103.594}}{\second{43.308}}
& \bgbs{105.216}{31.111}
\\

LoM + GMR
& --
& \bgbsfmt{\second{100.905}}{\second{100.905}}
& -- & -- & --
& \bgbs{105.955}{66.273}
& \bgbs{94.708}{53.427}
& \bgbs{99.441}{58.356}
& \bgbsfmt{\second{106.080}}{\second{43.838}}
& \bgbsfmt{\best{104.643}}{\best{55.622}}
& \bgbsfmt{\best{101.460}}{\best{56.082}}
& \bgbsfmt{\second{109.200}}{\second{44.864}}
\\

MotionCraft + GMR
& --
& \bgbs{88.715}{88.715}
& -- & -- & --
& \bgbsfmt{\best{113.184}}{\best{78.142}}
& \bgbsfmt{\best{109.449}}{\best{75.563}}
& \bgbsfmt{\best{111.383}}{\best{73.875}}
& \bgbsfmt{\best{112.396}}{\best{45.294}}
& \bgbs{110.487}{33.007}
& \bgbs{120.387}{41.832}
& \bgbsfmt{\best{111.205}}{\best{45.673}}
\\

TextOp
& --
& \bgbs{98.791}{98.791}
& -- & -- & --
& \bgbs{90.530}{47.566}
& \bgbs{87.797}{46.130}
& \bgbs{93.685}{46.839}
& \bgbs{94.975}{34.423}
& \bgbs{93.748}{39.790}
& \bgbs{99.116}{29.485}
& \bgbs{97.975}{34.137}
\\

UniAct
& --
& \bgbsfmt{\best{102.825}}{\best{102.825}}
& -- & -- & --
& \bgbs{101.909}{39.979}
& \bgbs{86.563}{55.304}
& \bgbs{93.481}{46.749}
& \bgbs{100.271}{34.326}
& \bgbsfmt{\second{103.783}}{\second{54.835}}
& \bgbs{104.827}{37.632}
& \bgbs{103.145}{35.544}
\\

UH-1
& --
& \bgbs{96.574}{96.574}
& -- & -- & --
& \bgbs{75.637}{46.138}
& \bgbs{74.760}{48.223}
& \bgbs{74.229}{48.111}
& \bgbs{105.816}{35.136}
& \bgbs{97.523}{39.349}
& \bgbs{102.715}{18.473}
& \bgbs{105.156}{35.047}
\\

\midrule

% ============================================================
% Video
% ============================================================
\rowcolor{videoyellow}
\multicolumn{13}{l}{\textbf{Video}}
\\

GEM + GMR
& -- & --
& \bgbsfmt{\second{119.884}}{\second{119.884}}
& -- & --
& \bgbsfmt{\second{117.741}}{\second{81.323}}
& \bgbsfmt{\second{116.026}}{\second{80.139}}
& \bgbsfmt{\second{118.187}}{\second{77.697}}
& \bgbsfmt{\second{118.979}}{\second{31.516}}
& \bgbsfmt{\second{104.582}}{\second{48.556}}
& \bgbsfmt{\best{103.435}}{\best{44.906}}
& --
\\

Video2Robot + GMR
& -- & --
& \bgbsfmt{\best{124.448}}{\best{124.448}}
& -- & --
& \bgbsfmt{\best{124.113}}{\best{119.886}}
& \bgbsfmt{\best{118.509}}{\best{114.479}}
& \bgbsfmt{\best{122.699}}{\best{118.751}}
& \bgbsfmt{\best{111.367}}{\best{37.834}}
& \bgbsfmt{\best{112.702}}{\best{48.588}}
& \bgbsfmt{\second{115.272}}{\second{44.075}}
& --
\\

\midrule

% ============================================================
% Trajectory
% ============================================================
\rowcolor{trajpurple}
\multicolumn{13}{l}{\textbf{Trajectory}}
\\

BFM-Zero
& -- & -- & -- & --
& \bgbsfmt{\best{95.387}}{\best{95.387}}
& -- & -- & -- & --
& -- & -- & --
\\

\midrule

% ============================================================
% Rhythm
% ============================================================
\rowcolor{rhythmred}
\multicolumn{13}{l}{\textbf{Rhythm}}
\\

LDA + GMR
& -- & -- & --
& \bgbsfmt{\second{94.009}}{\second{94.009}}
& --
& -- & -- & -- & --
& -- & -- & --
\\

MotionCraft
& -- & -- & --
& \bgbsfmt{\best{96.045}}{\best{96.045}}
& --
& -- & -- & -- & --
& -- & -- & --
\\

\bottomrule
\end{tabular}%
}

\endgroup

\caption{
Task-level Level~1 conditional-steering results on
\textbf{\textit{RoboSteer}}.
Each applicable entry is reported as \textit{BG/BS$_1$};
``--'' denotes an inapplicable task.
Video entries are aggregated over the human and skel subsets
using their actual evaluated sample counts as weights.
}
\label{tab:level1_tasks}
\end{sidewaystable}

\begin{sidewaystable}[p]
\centering
\begingroup
\scriptsize
\setlength{\tabcolsep}{3pt}
\renewcommand{\arraystretch}{1.15}
\arrayrulecolor{black}

\resizebox{\textwidth}{!}{%
\begin{tabular}{l|ccccccc}
\toprule
\multicolumn{1}{c|}{}
& \multicolumn{7}{c}{\textbf{Level 2: Constraint Steering}} \\
\cmidrule(lr){2-8}

\textbf{Model}
& \textbf{Spd.}
& \textbf{Amp.}
& \textbf{Dir.}
& \textbf{Order}
& \textbf{Times}
& \textbf{Traj.}
& \textbf{Body} \\
\midrule

\rowcolor{textblue}
\multicolumn{8}{l}{\textbf{Text}} \\

GEM + GMR
& 114.238 / 114.238 / 58.782
& \underline{113.930 / 113.930 / 74.166}
& 110.800 / 110.800 / 4.240
& 88.884 / 88.884 / 12.926
& 79.883 / 79.883 / 9.089
& 116.969 / 116.969 / 4.665
& 113.041 / 113.041 / 0.832 \\

LoM + GMR
& \underline{112.399 / 112.399 / 61.904}
& 111.484 / 111.484 / 59.152
& 115.237 / 115.237 / 8.729
& 77.758 / 77.758 / 17.178
& 69.542 / 69.542 / 3.128
& 111.097 / 111.097 / 18.310
& 109.484 / 109.484 / 1.612 \\

MotionCraft + GMR
& \textbf{122.370 / 122.370 / 65.919}
& \textbf{121.592 / 121.592 / 74.529}
& 123.280 / 123.280 / 14.730
& \textbf{82.230 / 82.230 / 20.754}
& \textbf{72.959 / 72.959 / 11.785}
& 116.091 / 116.091 / 10.486
& \textbf{118.997 / 118.997 / 11.714} \\

TextOp
& 93.554 / 93.554 / 46.777
& 92.072 / 92.072 / 53.257
& \textbf{97.408 / 97.408 / 24.362}
& \underline{73.113 / 73.113 / 18.674}
& \underline{70.040 / 70.040 / 10.336}
& \textbf{92.220 / 92.220 / 34.400}
& 87.240 / 87.240 / 0.268 \\

UniAct
& 102.444 / 102.444 / 51.222
& 103.360 / 103.360 / 58.287
& 98.011 / 98.011 / 0.000
& 77.910 / 77.910 / 9.604
& 72.658 / 72.658 / 7.409
& 101.142 / 101.142 / 0.000
& 105.630 / 105.630 / 6.802 \\

UH-1
& 106.350 / 106.350 / 59.324
& 105.126 / 105.126 / 62.128
& \underline{108.837 / 108.837 / 21.036}
& 86.741 / 86.741 / 13.849
& 78.444 / 78.444 / 3.971
& \underline{108.215 / 108.215 / 28.625}
& \underline{102.022 / 102.022 / 11.419} \\

\midrule

\rowcolor{audiogreen}
\multicolumn{8}{l}{\textbf{Audio}} \\

GEM + GMR
& 96.762 / 96.762 / 49.790
& \textbf{96.586 / 96.586 / 62.876}
& 93.886 / 93.886 / 3.593
& 89.239 / 89.239 / 12.978
& 80.562 / 80.562 / 9.166
& 99.077 / 99.077 / 3.951
& 95.627 / 95.627 / 0.704 \\

LoM + GMR
& \textbf{108.826 / 108.826 / 59.936}
& 107.951 / 107.951 / 57.277
& 110.889 / 110.889 / 8.400
& 78.031 / 78.031 / 17.194
& 74.839 / 74.839 / 3.366
& 107.853 / 107.853 / 17.775
& 105.773 / 105.773 / 1.557 \\

MotionCraft + GMR
& 89.636 / 89.636 / 48.285
& 89.153 / 89.153 / 54.646
& 87.406 / 87.406 / 10.444
& \textbf{82.299 / 82.299 / 20.772}
& \textbf{77.463 / 77.463 / 12.513}
& 91.584 / 91.584 / 8.272
& \underline{87.401 / 87.401 / 8.603} \\

TextOp
& 109.615 / 109.615 / 54.807
& \underline{108.332 / 108.332 / 62.663}
& \textbf{110.772 / 110.772 / 27.704}
& \underline{73.719 / 73.719 / 18.782}
& \underline{73.114 / 73.114 / 10.857}
& \textbf{111.564 / 111.564 / 41.616}
& 104.169 / 104.169 / 0.319 \\

UniAct
& 100.528 / 100.528 / 50.528
& 101.463 / 101.463 / 57.704
& 95.005 / 95.005 / 0.000
& 78.865 / 78.865 / 9.722
& 77.710 / 77.710 / 7.924
& 100.871 / 100.871 / 0.000
& 103.659 / 103.659 / 7.358 \\

UH-1
& \underline{104.982 / 104.982 / 58.561}
& 103.752 / 103.752 / 61.316
& \underline{106.443 / 106.443 / 20.574}
& 85.741 / 85.741 / 13.689
& 79.549 / 79.549 / 4.027
& \underline{106.865 / 106.865 / 28.268}
& \textbf{100.745 / 100.745 / 11.276} \\

\midrule

\rowcolor{videoyellow}
\multicolumn{8}{l}{\textbf{Video}} \\

GEM + GMR
& \underline{121.654 / 121.654 / 54.234}
& \textbf{121.015 / 121.015 / 61.248}
& \underline{121.898 / 121.898 / 5.410}
& \textbf{97.438 / 97.438 / 25.378}
& \underline{95.773 / 95.773 / 12.097}
& \underline{126.183 / 126.183 / 10.267}
& \underline{118.334 / 118.334 / 0.113} \\

Video2Robot + GMR
& \textbf{121.952 / 121.952 / 59.630}
& \underline{122.445 / 122.445 / 57.527}
& \textbf{116.332 / 116.332 / 25.971}
& \underline{95.395 / 95.395 / 23.116}
& \textbf{94.122 / 94.122 / 12.894}
& \textbf{121.527 / 121.527 / 24.434}
& \textbf{123.817 / 123.817 / 2.301} \\

\bottomrule
\end{tabular}%
}
\endgroup

\caption{Task-level evaluation results for Level~2 Constraint Steering in
\textbf{\textit{RoboSteer}}. Each applicable task reports BG, BS$_1$, and
BS$_2$ in the format ``BG / BS$_1$ / BS$_2$''; ``--'' denotes an
inapplicable task.}
\label{tab:level2_tasks}
\end{sidewaystable}

\end{document}